\documentclass[journal]{IEEEtran}

\usepackage[pdftex]{graphicx}
\usepackage{cite}
\usepackage{amsopn}
\usepackage{booktabs} 
\usepackage{url}
\usepackage{color}
\usepackage{multirow}
\usepackage{color}
\usepackage[switch]{lineno}
\usepackage{graphicx}
\usepackage{subfigure}
\usepackage{url}
\usepackage{epstopdf}
\usepackage{hyperref}
\usepackage[table,xcdraw]{xcolor}
\usepackage{amsmath, bm}
\usepackage{textcomp}
\usepackage{verbatim}

\begin{document}

\title{Contextualized Spatial-Temporal Network \\ for Taxi Origin-Destination Demand Prediction}

\author{Lingbo Liu,
        Zhilin Qiu,
        Guanbin Li,
        Qing Wang,
        Wanli Ouyang,
        and~Liang Lin

\thanks{
This work was supported in part by the National Science Foundation of China under Grant U1811463, Grant 61602533, and Grant 61702565, in part by the Fundamental Research Funds for the Central Universities under Grant 18lgpy63, in part by the Science and Technology Planning Project of Guangdong Province under Grant 2017B010116001, and in part by the SenseTime Research Fund.
(\emph{Corresponding author: Guanbin Li.})}

\thanks{L. Liu, Z. Qiu, G. Li, Q, Wang and L. Lin are with the School of Data and Cpmputer Science, Sun Yat-Sen University, Guangzhou 510006, China (e-mail: liulingb@mail2.sysu.edu.cn; qiuzhl3@mail2.sysu.edu.cn; liguanbin@mail.sysu.edu.cn; wangq79@mail.sysu.edu.cn; linliang@ieee.org).}
\thanks{W. Ouyang is with the School of Electrical and Information Engineering, the University of Sydney, Sydney, Camperdown, NSW, 2000 Australia (e-mail: wanli.ouyang@sydney.edu.au).}
}

\markboth{IEEE Transactions on Intelligent Transportation Systems}
{Liu \MakeLowercase{\textit{et al.}}: Taxi Origin-Destination Demand Prediction}

\maketitle


\begin{abstract}
Taxi demand prediction has recently attracted increasing research interest due to its huge potential application in large-scale intelligent transportation systems. However, most of the previous methods only considered the taxi demand prediction in origin regions, but neglected the modeling of the specific situation of the destination passengers. We believe it is suboptimal to preallocate the taxi into each region based solely on the taxi origin demand. In this paper, we present a challenging and worth-exploring task, called taxi origin-destination demand prediction, which aims at predicting the taxi demand between all region pairs in a future time interval. Its main challenges come from how to effectively capture the diverse contextual information to learn the demand patterns. We address this problem with a novel Contextualized Spatial-Temporal Network (CSTN), which consists of three components for the modeling of local spatial context (LSC), temporal evolution context (TEC) and global correlation context (GCC) respectively. Firstly, an LSC module utilizes two convolution neural networks to learn the local spatial dependencies of taxi demand respectively from the origin view and the destination view. Secondly, a TEC module incorporates both the local spatial features of taxi demand and the meteorological information to a Convolutional Long Short-term Memory Network (ConvLSTM) for the analysis of taxi demand evolution. Finally, a GCC module is applied to model the correlation between all regions by computing a global correlation feature as a weighted sum of all regional features, with the weights being calculated as the similarity between the corresponding region pairs. Extensive experiments and evaluations on a large-scale dataset well demonstrate the superiority of our CSTN over other compared methods for taxi origin-destination demand prediction.
%
\end{abstract}

\IEEEpeerreviewmaketitle
\section{Introduction}~\label{sec:introduction}
\IEEEPARstart{T}{axi} as one of the most common travel modes for urban residents, has greatly penetrated into people's daily life. Online taxicab requesting platforms, such as Didi Chuxing\footnote{\url{http://www.didichuxing.com/en/}}, Uber\footnote{\url{https://www.uber.com}} and Grab\footnote{\url{https://www.grab.com/}}, have recently experienced rapid expansion due to the convenience it brings to our daily travel. However, this huge industry still suffers from some inefficient operations~\cite{zhan2016graph} (i.e, long passenger waiting time~\cite{zhang2017taxi} and excessive vacant trips~\cite{yang2000macroscopic}). The main problem stems from the mismatch between supply and demand caused by inaccurate taxi demand prediction, which results in a large number of taxis gathering in some busy areas and causing oversupply, while in other remote areas the distribution of taxis was extremely sparse. The solution to this issue involves taxi demand prediction, which estimates the future taxi demand and helps to allocate the taxis to each region in advance.

\begin{table}
  \centering
  \caption{Comparison of the definition and scope of various related tasks.}
  \vspace{-2mm}
  \label{tab:different_task}
  \begin{tabular}{c|c}
    \hline
    Method & Task and Scope \\
    \hline\hline
    Zhang et al.\cite{zhang2017deep} & Traffic Inflow and Outflow Prediction \\
    Jin et al.\cite{jin2018spatio}    & {\color{red}in all regions} \\
    \hline
    Tong et al.\cite{tong2017simpler} & Taxi Demand Prediction\\
    Yao et al.\cite{yao2018deep}      & {\color{red}in all regions} \\
    \hline
    Toqu et al.\cite{toque2016forecasting} & Traffic Flow or Demand Prediction\\
    Azzouni et al.\cite{azzouni2017long}   & {\color{red}between some well-designed positions} \\
    Yang et al.\cite{yang2017daily}        & (e.g., highway toll booths, subway and bus stations) \\
    \hline
    \multirow{2}{*}{Zhou et al.\cite{zhou2018predicting}} & Passenger Pickup/Dropoff Demand Prediction \\
    & {\color{red}in all regions} \\
    \hline
    \multirow{2}{*}{Ours} & Taxi Demand Prediction\\
    & {\color{red}between all regions} \\
    \hline
\end{tabular}
\end{table}

As a crucial task in intelligent transportation systems (ITS), taxi demand prediction has attracted a wide range of research interest and achieved notable successes~\cite{moreira2013predicting,qian2017forecasting,ke2017short,tong2017simpler,yao2018deep}. However, most of the existing methods only model the demand of the taxi at the departure place and estimate the requests for taxis in all regions or some specific locations, ignoring the influence of the passenger destination. We believe the information of passengers' destinations is critical for the taxi preallocation systems. Without considering the distribution of passengers destinations, the taxi preallocation systems deploy the taxis in advance based solely on the predicted taxi origin demand, which may suffer from the following issues:
\begin{itemize}
\item Limited by the city management rules (such as the driving restriction policy\footnote{\url{http://zhengce.beijing.gov.cn/library/192/33/50/438650/1552930/index.html}} in Beijing), some drivers are only allowed to drive in some specified regions. If a taxi driver is assigned to a region where most passengers are to go to a place where the driver is restricted, he/she cannot take orders, which may result in waste of resources.
\item Some drivers prefer to carry passengers in their familiar regions. Meanwhile, some drivers are unwilling to take the short trip orders for little profit. If the destinations of most passengers in the driver's preallocated region are out of his/her operating regions or too close to the pickup locations, the driver may reject those requests.
\item If a driver is dispatched to a region where most passengers will travel to his/her unfamiliar regions, the driver may spend more time to carry the passengers to their destination, even though guided by GPS navigation. This will reduce the taxi market operating efficiency and the levels of passenger satisfaction.
\end{itemize}
In literature, some works~\cite{toque2016forecasting,azzouni2017long,yang2017daily} have been proposed to estimate the traffic flow or demand between some well-designed positions, such as highway toll booths, subway and bus stations. However, taxi passengers can be anywhere and these traffic flow estimation system for limited positions may not be suitable for citywide taxi preallocation. Therefore it becomes desirable to predict the taxi demand between every two regions and optimize the taxi allocation mechanism.

In this paper, we propose a challenging taxi origin-destination demand prediction task, which aims to predict the future taxi demands between any two regions. If the taxi origin demand and the destinations of passengers are well predicted, we can preallocate the taxi more efficiently to meet the passengers' requests and simultaneously avoid all above issues.
The key challenges of the proposed task lie in how to capture the diverse spatial-temporal contextual information to learn the demand patterns. For example, some regions that are spatially adjacent usually have the similar demand patterns (e.g, the number of taxi requests and the demand trends), which is called as local spatial context (LSC) in our work. Moreover, even though two regions are spatially distant, the demand patterns may still have some relevance, if they share similar functionality (e.g., both of them are residential districts). We call this relationship between two far-apart regions as global correlation context (GCC). Finally, taxi demand is a time-varying process and its evolution is related to various factors, such as its current states and the ever-changing meteorology, which is formulated as temporal evolution context (TEC).

Recently, deep neural networks have facilitated great advances in context modeling~\cite{liang2015human,zhao2015saliency,li2016lstm}. Inspired by this, we address the problem of taxi origin-destination demand prediction with a novel Contextualized Spatial-Temporal Network (CSTN), which well integrates the local spatial context, temporal evolution context, and global correlation context into a unified framework. Specifically, our proposed network consists of three components, including a LSC module, a TEC module and a GCC module, respectively for the three types of context modeling.
Firstly, a LSC module utilizes two convolution neural networks to learn the local spatial dependencies of taxi demand respectively from the origin view and the destination view. The output of the two networks would be combined to generate the final local spatial feature, which involves the hybrid information of taxi demand patterns from different views. Secondly, a TEC module incorporates both the local spatial features of taxi demand and the meteorological information to a CNN-LSTM network~\cite{xingjian2015convolutional} (convolutional long short-term memory network) for the analysis of taxi demand evolution. Thirdly, to capture the correlation between the far-apart regions, the GCC module computes the similarity between any two regions and generates the global correlation feature of each region by summing the features of all regions with the similarity weights. In this way, each region contains the information of all regions and it is mainly relevant to the regions that have high similarities with it. Finally, we integrate the local spatial-temporal feature generated by TEC module and the global correlation feature generated by GCC module to predict the future taxi origin-destination demand.

The main contributions of this work are three-fold:
\begin{itemize}
\item We extend the existing taxi demand prediction to the task of taxi original-destination demand prediction, which is more worth-exploring for intelligent transportation systems. To the best of our knowledge, we are the first to study the interregional taxi demand prediction.

\item We propose a novel Contextualized Spatial-Temporal Network to address this task, which well integrates the local spatial context, temporal evolution context and global correlation context into a unified framework.

 \item Extensive experiments on a large-scale benchmark of taxi original-destination demand prediction demonstrate that our approach outperforms existing state-of-the-art methods by a margin.

\end{itemize}

The rest of the paper is organized as follows. We firstly review some related works in Section~\ref{sec:related_works} and define the taxi original-destination demand problem with some notations in Section~\ref{sec:definition}. We then introduce the proposed Contextualized Spatial-Temporal Network in Section~\ref{sec:model} and conduct extensive experiments in Section~\ref{sec:experiments}. Finally we conclude this paper in Section~\ref{sec:conclusion}.

\section{Related Works}~\label{sec:related_works}
In this paper, we utilize deep neural networks to forecast the interregional taxi demand, which is closely related to the \textit{taxi demand prediction} and \textit{origin-destination estimation}. We will thoroughly review the relevant works of these two categories of researches in the following subsections.

\subsection{Taxi Demand Prediction}
Due to its huge potential application in ITS, taxi demand prediction has been extensively studied~\cite{sun2004querying,anwar2013changinow,zhang2016framework,qiu2019taxi,xu2017real}.
Moreira-Matias et al.~\cite{moreira2013predicting} proposed to aggregate the GPS signals into histogram time series and applied them to predict the demand with a Poisson Model and an AutoRegressive Moving Average model.
Yuan et al.~\cite{yuan2011find} presented a recommender to provide taxi drivers accurate locations to pick up passengers quickly with historical GPS trajectories of taxicabs.
Li et al.\cite{li2012prediction} forecast the spatio-temporal variations of passengers at the given hotspot with an improved ARIMA-based prediction model.
All above methods require the taxi trajectories. The trajectory-free prediction has recently attracted increasing attention. A pioneer work was proposed by Tong et al.~\cite{tong2017simpler}, in which they utilized the taxi-calling records from some online taxicab requesting platforms to predict the taxi demand with a unified linear regression model.

Recently, the success of deep learning on various computer vision tasks~\cite{chen2016disc,liu2018crowd,li2018contrast,ouyang2018jointly,qiu2019crowd,liu2019facial} motivates researchers to adopt the deep neural network to handle this task.
Wang et al.\cite{wang2017deepsd} designed a neural network framework using context data from multiple sources to predict the gap between taxi supply and demand.
Xu et al.\cite{xu2017real} proposed a sequence learning model that can predict future taxi requests in each area of a city based on the recent demand and other relevant information.
Yao et al.\cite{yao2018deep} proposed a Deep Multi-View Spatial-Temporal Network framework to model both spatial and temporal relations of taxi demand.
Rodrigues et al.\cite{rodrigues2019combining} combined time-series and textual data to  forecast the taxi demand in event areas with two hybrid deep learning architectures.
Recently, Zhou et al.\cite{zhou2018predicting} built an attention-based neural network to predict the passenger pickup/dropoff demand on each region, but they were still not applicable to taxi demand prediction between region pairs.

All the above methods only forecast the taxi demand per unit time in each region or at some specific locations.
In contrast, our method attempts to predict interregional taxi demand, which can help taxi preallocation systems to allocate the taxis more efficiently. Moreover, our proposed CSTN explicitly captures the local spatial context, temporal evolution context and global correlation context in one united framework to infer more accurate taxi demand patterns.

\begin{figure}
\centerline{\includegraphics[width=.48\textwidth]{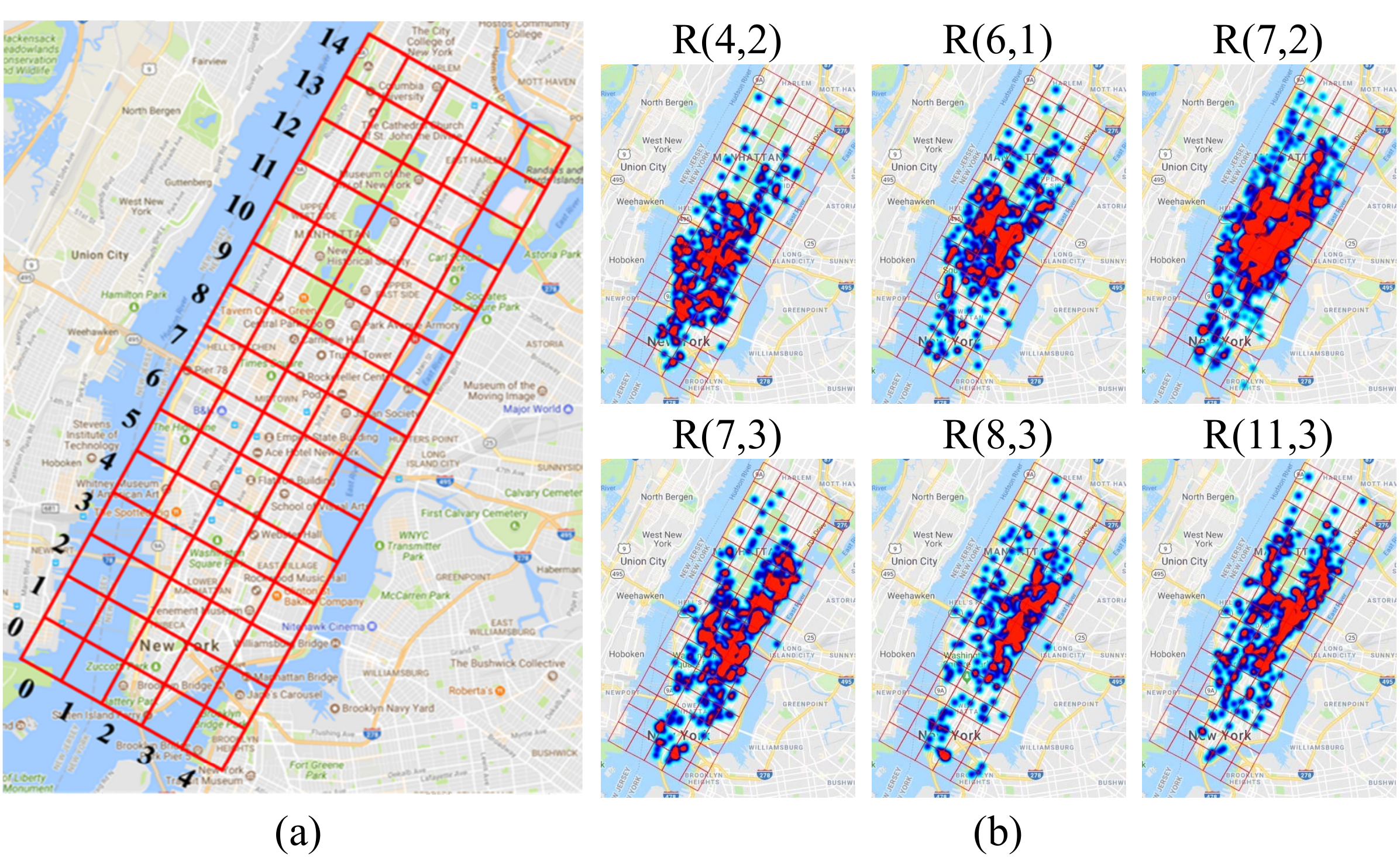}}
\vspace{-4mm}
\caption{\textbf{(a)} Illustration of the region partition on a city. We partition a city into a grid map based on the longitude and latitude. Here is the example of the Manhattan in New York City. \textbf{(b)} Visualization of the taxi demand from origin view by mapping the passengers' pick-up locations back to the geo-coordinates on Google map. The sub-figure with title ``R(i,j)'' is the taxi demand from all regions to the region R(i, j) during 8:00-8:30 am, May. 8, 2014.}
\vspace{-2mm}
\label{fig:NYC_map}
\end{figure}

\subsection{Origin-Destination Estimation}
Origin-Destination Estimation~\cite{tamin1989transport,cascetta1988unified,zhou2006dynamic} aims to estimate the flow between the endpoints of the studied traffic network, given the flow count and other observations of several traffic links.
Existing research works on this task can be divided into two categories, including static estimation and dynamic estimation.
The static approaches~\cite{tamin2003development,hazelton2003some} consider the traffic flow as time-independent and estimate the average demand, which are suitable for long-time transportation planning and design purpose. On the other hand, dynamic approaches~\cite{zhou2003dynamic,hazelton2008statistical} estimate the time-variant flow between each origin and destination, which can be used for short time route guidance and dynamic traffic assignment.
These works generally take as input the flow count of some links collected from well-designed positions (e.g., highway toll booths, the intersection of main street, express road, subway and bus stations) and some prior information~(e.g. the proportion of different origin-destination pair).
When considering a huge number of positions, the OD matrix would become high-dimensional and hard to be computed.
Some previous studies~\cite{yang2017daily,8113466} attempted to resolve this issue through dimension reduction technology.

However, these methods are designed to estimate the traffic flow between some specific positions and are not effective for citywide taxi preallocation, as taxi passengers can be located in any area. In contrast, our method divides a city into multiple regions and forecasts the taxi demand between these regions.

\section{Preliminaries}~\label{sec:definition}
In this section, we first define some notations and then formulate the taxi origin-destination demand prediction problem based on these notations.

\begin{figure}
\centerline{\includegraphics[width=0.48\textwidth, height=2.8cm]{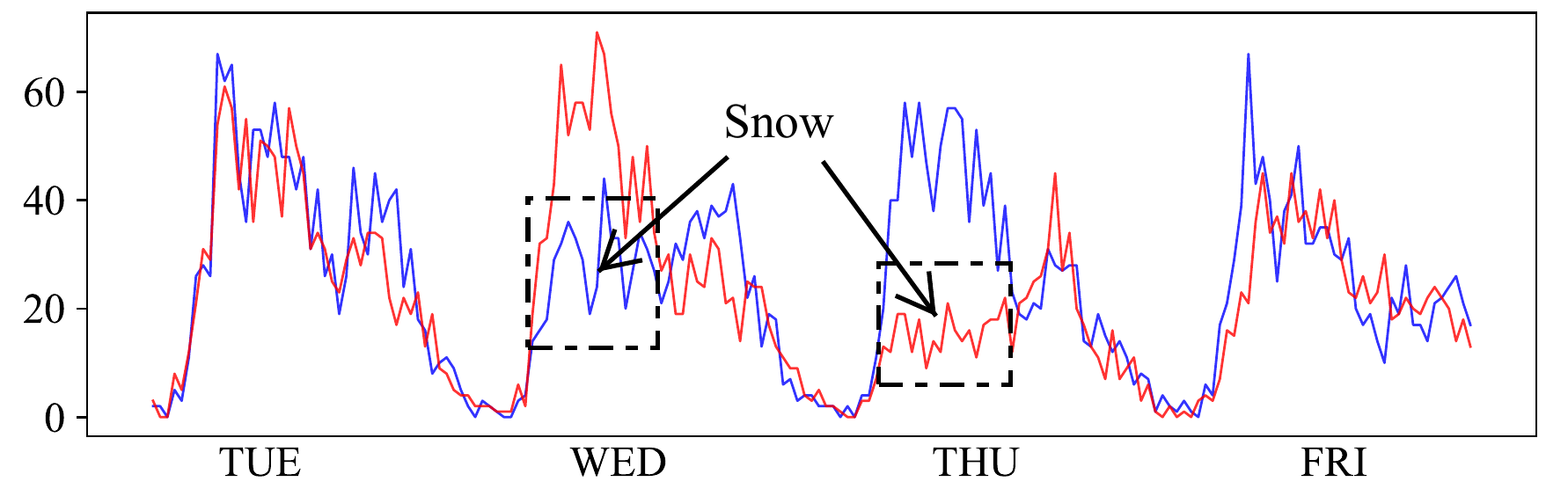}}
\vspace{-4mm}
\caption{Influence of meteorological conditions on taxi demand. We show the taxi demand from region(5, 2) to region (7, 2) of New York City in two time periods: Feb 4-7 ({\color{red}Red}) and Feb 11-14 ({\color{blue}Blue}) in 2014. We can see that the heavy snow sharply reduces the taxi demand compared to the same day of the adjacent week.}
\vspace{-0mm}
\label{fig:OD-weather}
\end{figure}

\begin{figure*}[t]
\begin{center}
 \includegraphics[width=1.60\columnwidth]{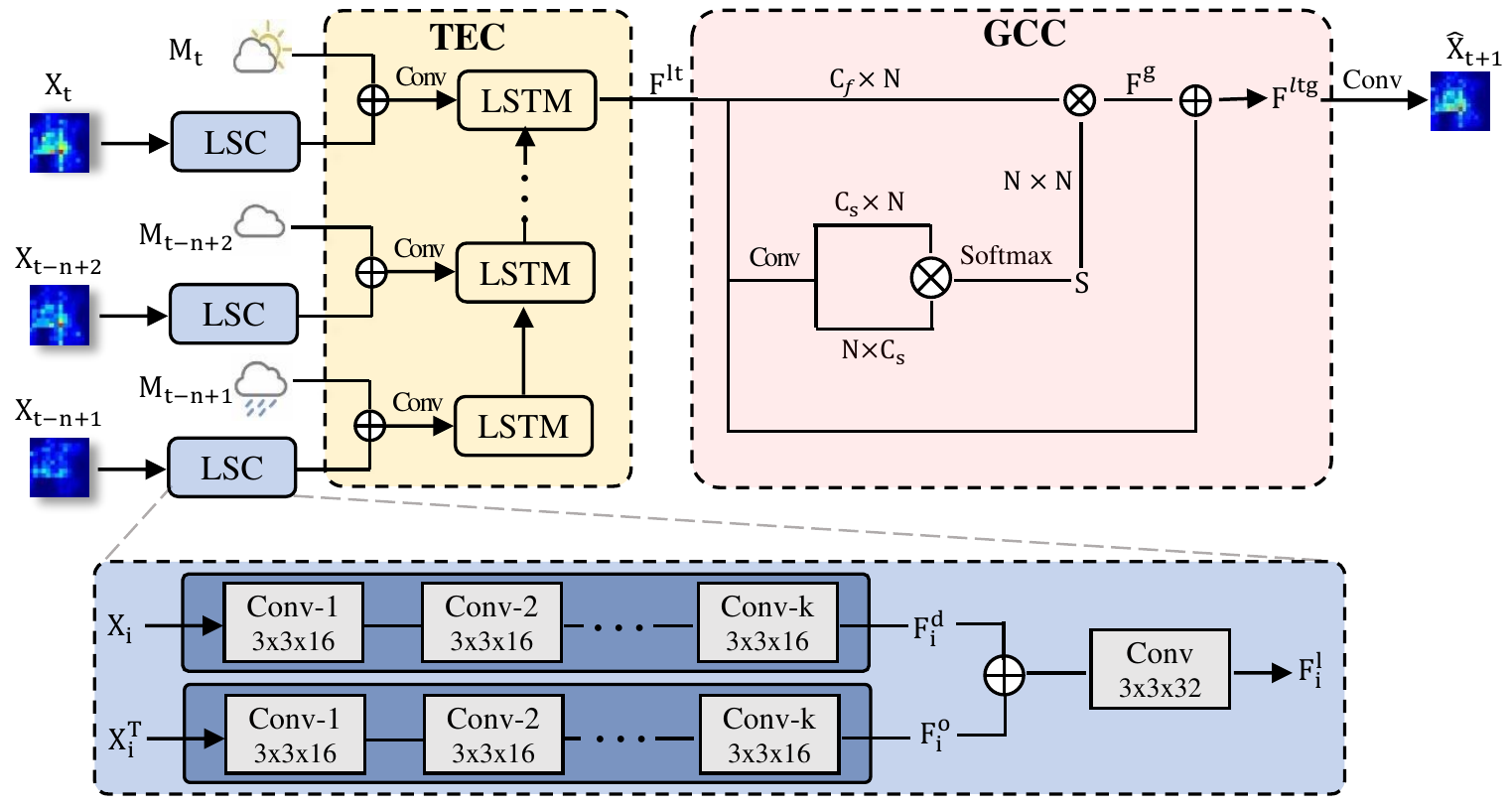}
 \vspace{-6mm}
\end{center}
  \caption{The architecture of the proposed Contextualized Spatial-Temporal Network (CSTN) for taxi origin-destination demand prediction.
  ${X_i}$ denotes the OD matrix in time interval ${i}$, while ${X_i^T}$ is the DO matrix called in our work. ${M_i}$ is the meteorological data. $C_{lt}$ is the channel number of feature $F^{lt}$ and $C_s$ is the channel number of $F_{s}$ . $N$ is the total number of the regions.
  ``$\bigoplus$'' denotes feature concatenation and ``$\bigotimes$'' refers to the dot product operation.
  CSTN consists of three components for three types of context modeling respectively.
  The LSC module feeds ${X_i}$ and ${X_i^T}$ into a Two-View ConvNet to respectively learn the local spatial context from the origin view and destination view and then combines the output of the two ConvNet. The TEC module recurrently takes the local feature ${F_i^l}$ generated by LSC and the meteorological data $M_i$ to learn the temporal evolution context of the taxi demand with a ConvLSTM. The GCC module computes the similarity between all regions and generates the global correlation feature of each region by summing the features of all regions with the similarity weights.}
\vspace{-2mm}
\label{fig:network-structure}
\end{figure*}

\textbf{Region Partition:}
In this work, we focus on the taxi origin-destination demand prediction between regions, rather than the specific positions. There are many ways to divide a city into multiple regions in terms of different granularities and semantic meanings, such as road network~\cite{deng2016latent} and zip code tabular~\cite{qian2017forecasting}. Inspired by the previous works~\cite{zhang2017deep,yao2018deep}, we partition a city into ${H{\times}W}$ non-overlapping grid map based on the longitude and latitude. Each rectangular grid represents a different geographical region in the city. The region on the ${i^{th}}$ row and ${j^{th}}$ column of the grid map is denoted as $R(i, j)$ in the following sections. Figure~\ref{fig:NYC_map}(a) illustrates the partitioned regions of the Manhattan in New York City. With this simple partition method, the raw taxi request records could be directly transformed into matrix or tensor, which is the most common format of input data of the deep neural networks.

\textbf{Taxi Origin-Destination Demand}:
In taxi calling industry, the taxicab companies or online platforms, such as Didi Chuxing and Uber, would receive a large number of taxi requests from passengers every second. Each raw taxi request contains the origin location, destination location, timestamp and other information (e.g., user identification and phone number) of the passengers. In our work, the taxi origin-destination demand is defined as the total number of taxi requests from the origin region to the destination region in each time interval.

We denote the taxi origin-destination demand in time interval $t$ as a 3D matrix ${\bm{X}_t} \in R^{N{\times}H{\times}W}$, where ${H}$ and ${W}$ are the height and width of the city grid map respectively. ${N}$ is the total number of the regions in city and it is equal to ${H{\bm\cdot}W}$. Specifically, ${\bm{X}_t(d, i_o, j_o)}$, in which the destination index ${d}$ is equal to ${W{\bm\cdot}i_d+j_d}$, is the demand from origin region $R(i_o, j_o)$ to destination region $R(i_d, j_d)$, The value of ${\bm{X}_t(d, i_o, j_o)}$ can be measured from the taxi request records in time interval $t$. In particular, the ${d^{th}}$ channel of ${\bm{X}_t}$, denoted as ${\bm{X}_t(d) \in R^{H{\times}W}}$, is the taxi demand from all regions to region $R(i_d, j_d)$. Figure~\ref{fig:NYC_map}(b) shows some channels of ${\bm{X}_t}$ by mapping the passengers' pick-up locations back to the geo-coordinate on Google map. The taxi origin demand, denoted as ${\bm{O}_t \in R^{H{\times}W}}$, can be easily calculated by ${\sum_{d=0}^{N-1} \bm{X}_t(d)}$.

\textbf{Taxi Origin-Destination Demand Prediction}:
The taxi origin-destination demand prediction problem in our work aims to predict the taxi origin-destination demand $\bm{X}_{t}$ in time interval $t$, given the data until time interval $t-1$. As shown in Figure~\ref{fig:OD-weather}, the taxi demand is seriously affected by the meteorological conditions, so we also incorporate the historical meteorological data to handle this task and we denote the meteorological data in time interval $i$ as ${\bm{M}_i}$. The collection and preprocessing of taxi demand data and meteorological data are described in Section~\ref{sec:data}. Therefore, our final goal is to predict $X_t$ with the historical demand data $\{ \bm{X}_i | i = t-n+1,...,t\}$ and meteorological data $\{\bm{M}_i | i = t-n+1,...,t\}$, where ${n}$ is the sequence length of time intervals.

\section{The Proposed Method}~\label{sec:model}
In this section, we propose a novel Contextualized Spatial-Temporal Network (CSTN) for taxi origin-destination demand prediction.
As shown in Figure~\ref{fig:network-structure}, our network consists of three components for three types of context modeling respectively.
First, the LSC module utilizes two convolutional neural networks to learn the local spatial context of taxi demand from origin view and destination view.
Second, the TEC module incorporates both the local spatial features of taxi demand and the meteorological information to a ConvLSTM for the analysis of taxi demand evolution.
Third, the GCC module generates the global correlation feature of each region by summing the features of all regions with the calculated similarity weights.

\subsection{Local Spatial Context Modeling}~\label{sec:LSC}
Generally, the taxi demand is usually related to local spatial location, and the spatially adjacent regions may have the similar demand patterns.
For instance, people tend to depart from residence regions and head to employment regions in the morning rush hours.
In this case, most of the residence regions in city suburb have high origin demands, while most of the working area in city center have high destination demands. Vice versa in the evening rush hours.
Recently, Yao et al.~\cite{yao2018deep} modeled the local spatial context of taxi origin demand with convolutional layers, but they neglected the context of destination demand.

\begin{figure}
\centerline{\includegraphics[width=.50\textwidth]{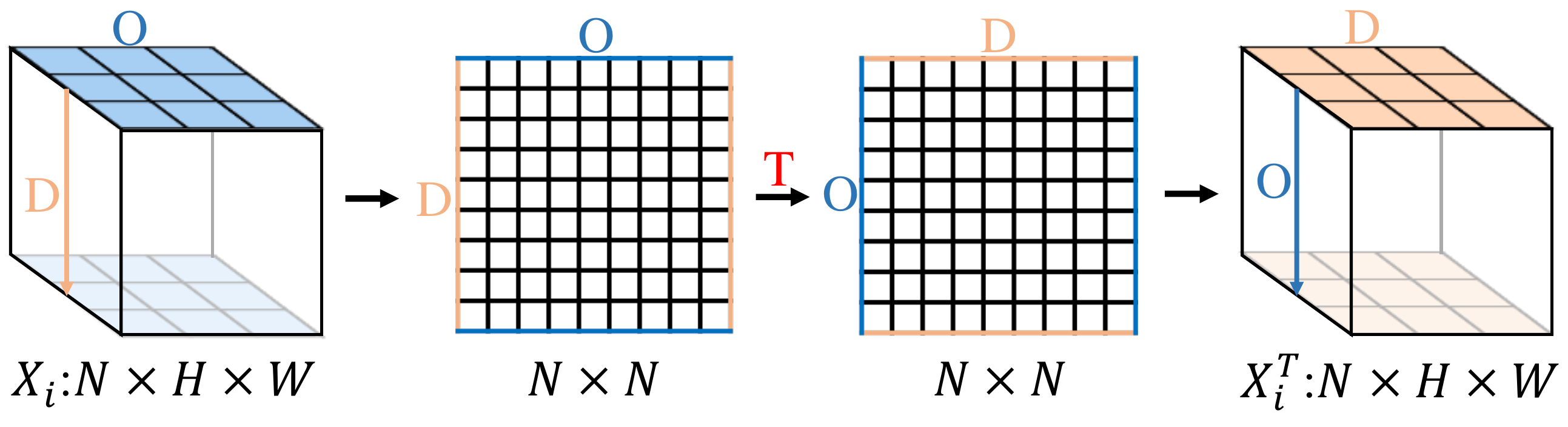}}
\vspace{-2mm}
\caption{Illustration of the generation process of DO matrix from OD matrix. $N$ is the total number of regions and it is equal to $H{\times}W$. ${T}$ denotes the matrix transposition. Each channel of the OD matrix ${X_i}$ is the taxi demand from all origin regions to the corresponding region. ${X_i^T}$ is called the DO matrix in our work and each channel denotes the taxi demand from the corresponding region to all destination regions.}
\vspace{-2mm}
\label{fig:DO}
\end{figure}

In this work, our proposed LSC module simultaneously captures the local spatial context of taxi demand from both the origin view and destination view. As described in Section~\ref{sec:definition}, each channel of the OD matrix ${\bm{X}_i}$ is the taxi demand from all origin regions to the corresponding region, thus we define the convolution operations on ${\bm{X}_i}$ as origin view modeling.
To model the local spatial context from destination view, we generate a DO matrix ${\bm{X}_i^T}$ from ${\bm{X}_i}$ with the transformation process described in Figure~\ref{fig:DO}. Specifically, we first reshape ${\bm{X}_i}$ to be a 2D matrix and then conduct the common transposition operation. Finally, the transpose matrix is reorganize to be a 3D tensor ${\bm{X}_i^T}$. Each channel of ${\bm{X}_i^T}$ is the taxi demand from the corresponding region to all destination regions.

As shown in the bottom of Figure~\ref{fig:network-structure}, our LSC module is implemented by a Two-View ConvNet, which takes ${\bm{X}_i}$ and ${\bm{X}_i^T}$ as input to respectively capture the local spatial context from different views. The origin view CNN contains ${K}$ convolutional layers. Each convolutional layer has 16 filters of kernel size of $3\times3$, followed by a Rectified Linear Unit (ReLU). To maintain the same resolution in space, the strides of all convolutional layers are set to 1 and no pooling layers are adopted in the network.
The destination view CNN has the same network structure with the origin view CNN.
In time interval ${t}$, the origin view CNN takes ${\bm{X}_i}$ as input and its output feature ${\bm{F}_i^o}$ only contains the local spatial context of origin view.
Meanwhile, the destination view CNN takes ${\bm{X}_i^T}$ as input and its output feature ${\bm{F}_i^d}$ only contains the local spatial context of destination view.
To capture the integrated local spatial context, we finally fuse these two features using a convolutional layer with 32 filters. The whole pipeline of our LSC module can be expressed as:
\begin{equation}
\begin{split}
\bm{F}_i^o = &\textbf{CNN}(\bm{X}_i, \bm{w}_o), \\
\bm{F}_i^d = &\textbf{CNN}(\bm{X}_i^T, \bm{w}_d), \\
\bm{F}_i^l = &\textbf{Conv}(\bm{F}_i^o \oplus \bm{F}_i^d, \bm{w}_{od}),
\end{split}
\end{equation}%
where $\bm{w}_o$ and $\bm{w}_d$ are the parameters of the origin view CNN and destination view CNN respectively. $\mathbf{w}_{od}$ denotes the parameters of the fusion convolutional layer and $\oplus$ denotes the feature concatenation operation. ${\bm{F}_i^l}$ is the final local spatial feature, which contains the local spatial context of taxi demand from both the origin view and the destination view.

\subsection{Temporal Evolution Context Modeling}
Taxi demand is a time-varying process and it is usually affected by diverse complicated factors. Besides its own internal states, the meteorological conditions also impact the future demand.
For instance, a sustained snowfall may seriously weaken the travel willingness of residents and cause a decrease in taxi demand, as shown in Figure~\ref{fig:OD-weather}.
Therefore, we incorporate the historical demand feature and the ever-changing meteorological conditions to grasp the evolving tendency of taxi demand along the temporal dimension.

Fully Connected Long Short-term Memory Network (FC-LSTM)\cite{hochreiter1997long} has been proven to be powerful for temporal context modeling, but it fails to preserve the local spatial context captured by the aforementioned Two-View ConvNet.
In this work, we model the temporal evolution context of taxi demand with an advanced Convolutional LSTM (ConvLSTM)~\cite{xingjian2015convolutional}.
Compared with FC-LSTM, ConvLSTM can preserve the structural locality of input feature, with the convolutional structures in both the input-to-state and state-to-state connections.
Moreover, it can effectively accumulate the previous sequential information by maintaining a memory cell.
Specifically, at iteration ${k}$, given the input $\mathcal{\bm{X}}_k$, the ConvLSTM updates its memory cell $\bm{c}_k$ with an input gate $\bm{i}_k$ and a forget gate ${\bm{f}_k}$, and controls its hidden state ${\bm{h}_k}$ with an output gate ${\bm{o}_k}$. Its formulation can be expressed as follows:
{
\small
\begin{equation}
\begin{split}
\bm{i}_{k}=&\sigma \left ( \bm{w}_{xi}\ast \mathcal{\bm{X}}_{k}+\bm{w}_{hi}\ast \bm{h}_{k-1}+\bm{w}_{ci}\circ \bm{c}_{k-1}+ b_{i} \right )\\
\bm{f}_{k}=&\sigma \left ( \bm{w}_{xf}\ast \mathcal{\bm{X}}_{k}+\bm{w}_{hf}\ast \bm{h}_{k-1}+\bm{w}_{cf}\circ \bm{c}_{k-1}+ \bm{b}_{f} \right )\\
\bm{c}_{k}=&\bm{f}_{i}\circ \bm{c}_{k-1}+\bm{i}_{i}\circ tanh \left ( \bm{w}_{xc}\ast \mathcal{\bm{X}}_{k}+\bm{w}_{hc}\ast \bm{h}_{k-1}+ b_{c} \right )\\
\bm{o}_{k}=&\sigma \left ( \bm{w}_{xo}\ast \mathcal{\bm{X}}_{k}+\bm{w}_{ho}\ast \bm{h}_{k-1}+\bm{w}_{co}\circ \bm{c}_{k}+ b_{0} \right )\\
\bm{h}_{k}=&\bm{o}_{k}\circ tanh\left ( \bm{c}_{k} \right )
\end{split}
\label{lstm}
\end{equation}
}%
where ${\circ}$ denotes the Hadamard product, and ${\sigma}$ is the logistic sigmoid function. Symbol ${\ast}$ denotes the convolutional operator and $\bm{w}_{\alpha \beta}\left (\alpha \in\left \{ x,h,c\right \} ,\beta\in\left \{ i,f,o,c\right \}\right )$ are the parameters of convolutional layers in ConvLSTM.

We aim to predict the taxi demand $\bm{X}_t$ with the historical demand and the meteorological conditions of previous ${n}$ time intervals.
For the meteorological data ${\bm{M}_i}$, we encode it with a Multiple Layer Perceptron (MLP), which is implemented by three stacked fully-connected layers with 64, 16 and 8 neurons respectively. Then we copy the output feature of the MLP ${H{\bm\cdot}W}$ times and construct a 3D meteorological feature ${\bm{F}_i^m \in R^{8{\times}H{\times}W}}$. We combine ${\bm{F}_i^l}$ and ${\bm{F}_i^m}$ with a convolutional layer, which is expressed as:
{
\begin{equation}
\bm{F}_i^{lm} = \textbf{Conv}(\bm{F}_i^l \oplus \bm{F}_i^m, \bm{w}_{lm}),
\end{equation}}%
where $\oplus$ is the feature concatenation operation and $\bm{w}_{lm}$ denotes the parameters of the convolutional layer with 32 filters.
${\bm{F}_i^{lm}}$ is the local spatial feature that integrates the meteorological information.

We feed the features ${\bm{F}_{t-n+1}^{lm}, \bm{F}_{t-n+2}^{lm}, ..., \bm{F}_{t}^{lm}}$ into the ConvLSTM sequentially. At iteration $i$, the ConvLSTM takes ${\bm{F}_{t-n+i}^{lm}}$ as input and accumulates the previous sequential information to the memory cell $\bm{c}_i$ with Eq.(\ref{lstm}). After ${n}$ iteration, the hidden state of ConvLSTM is denoted as ${\bm{h}_n}$. We generate the local spatial-temporal feature $\bm{F}^{lt}$ by feeding ${\bm{h}_n}$ into a convolutional layer with ${C_{lt}}$ filters, which is expressed as:
\begin{equation}
\bm{F}^{lt} = \textbf{Conv}(\bm{h}_n, \bm{w}_{lt}),
\end{equation}%
where $\bm{w}_{lt}$ is the parameters of the convolutional layer and $\bm{F}^{lt}$ encodes the temporal evolution context of the taxi demand.

\subsection{Global Correlation Context Modeling}~\label{sec:GCC}
In the above two modules, the ConvNets and ConvLSTM only capture and maintain the local context of taxi demand.
However, the taxi demand distribution is also related to the attribute of the regions, e.g., most of the residential regions in different areas of the city may have high taxi demands in the morning rush hours. Therefore, even if the two regions are far apart in distance, they may still have similar taxi demand patterns as long as the attributes of the two regions are consistent. We call this kind of correlation as global correlation context.

Inspired by the recent work~\cite{wang2017non}, we capture the global correlation between all regions with a global feature fusion operation. Specifically, we generate the global correlation feature of each region as a weighted sum of all regional features, with the weights being calculated as the similarity between the corresponding region pairs. In this way, each region contains the information of all regions and it is mainly relevant to the regions of high similarities with it.

We detail each step of our GCC module as follows. Firstly, we feed $\bm{F}^{lt}$ into a convolutional layer with ${C_s}$ filters to generate an embedded feature ${\bm{F}_s}$ and then reshape it into a 2D matrix, which can be expressed as:
\begin{equation}
\begin{split}
\bm{F}_s = & \textbf{Conv}(\bm{F}^{lt}, \bm{w}_{s}),  \\
\bm{F}_s:{~}& R^{C_s \times H \times W } \rightarrow R^{C_s \times N}, \\
\end{split}
\end{equation}%
where $N$ is equal to $H{\bm\cdot}W$ and $\bm{w}_{s}$ is the convolutional parameters. Each column of $\bm{F}_s$ stands for the feature of a region. We further calculate the similarity matrix $\bm{S} \in R^{N \times N}$ as a dot-product operation between ${\bm{F}_s \in R^{C_s \times N}}$ and its transposed matrix ${\bm{F}_s^T \in R^{N \times C_s}}$, and perform the Softmax operation on each column of $\bm{S}$, which is expressed as:
\begin{equation}
\begin{split}
\bm{S} = \textbf{Softmax}({\bm{F}_s^T} \otimes \bm{F}_s),  \\
\end{split}
\end{equation}%
where $\otimes$ denotes the dot product operation. $\bm{S}_{i,j}$ is the normalized similarity weight between the two regions with index ${i}$ and index ${j}$.

After obtaining the similarity matrix $\bm{S}$, we compute the global correlation feature of each region by summing the features of all regions with the calculated similarity weights. We implement this process with a dot-product operation. We reshape $\bm{F}^{lt}$ to dimension ${C_{lt} \times N}$ and then dot-product $\bm{F}^{lt}$ and $\bm{S}$ to compute the global feature $\bm{F}^g$, which is further reshaped to dimension ${C_{lt} \times H \times \ W}$. The entire process can be expressed as :
\begin{equation}
\begin{split}
\bm{F}^{lt}:{~}& R^{C_{lt} \times H \times W } \rightarrow R^{C_{lt} \times N}, \\
\bm{F}^g = &F^{lt} \otimes  S,  \\
\bm{F}^g:{~}& R^{C_{lt} \times N  } \rightarrow R^{C_{lt} \times H \times W}
\end{split}
\end{equation}%
The feature $\bm{F}^g \in R^{C_{lt} \times H \times W}$ encodes the global correlation context, but lacks of structural locality, which would cause performance degradation. Therefore, we generate a new feature ${\bm{F}^{ltg}}$ by concatenating ${\bm{F}^{lt}}$ and $\bm{F}^g$. The feature ${\bm{F}^{ltg}}$ is thus incorporated with hybrid information of the local spatial context, temporal evolution context and global correlation context.

Finally, we predict the taxi origin-destination demand in time interval ${t}$, denoted as ${\hat{\bm{X}}_{t+1} \in R^{N{\times}H{\times}W}}$, by feeding ${\bm{F}^{ltg}}$ into a linear regression, which can be formulated as:
\begin{equation}
\label{equ:forecast}
\hat{\bm{X}}_{t+1} = tanh(\mathcal T(\bm{F}^{ltg})),
\end{equation}%
where ${\mathcal T}$ is the linear regression implemented by a convolutional layer with ${N}$ filters and the hyperbolic tangent ${tan}$ ensures the output values are between -1 and 1\footnote{When training, we use Min-Max linear normalization method to scale the origin-destination demand matrices into the range ${[-1, 1]}$. We re-scale the predicted values back to the normal values and then compare with the ground truth while performing evaluation.}.

\subsection{Implementation Details}
We implement our Contextualized Spatial-Temporal Network with Tensorflow~\cite{abadi2016tensorflow}. In LSC module, the layer number ${K}$ is set to 3, which means each ConvNet consists of three convolutional layers. In TEC module, all convolutional layers in ConvLSTM have 32 filters and the channel number ${C_{lt}}$ of feature ${F^{lt}}$ is set to 75. In GCC module, the channel number ${C_{s}}$ of feature ${F_{s}}$ is set to 64.
For the whole model, the filter parameters of all convolutional layers and the fully-connected layers are initialized by Xavier~\cite{glorot2010understanding}. The size of a minibatch is set to 64. The learning rate is initially set to $10^{-4}$ and multiplied by 0.1 every 200 epochs.
We optimize our network in an end-to-end manner via Adam optimization~\cite{kingma2014adam} by minimizing the Euclidean loss between the ground truth and the predicted result. It takes 7 hours to train our network for 700 epochs with an NVIDIA K80 GPU.

\section{Experiments}~\label{sec:experiments}
In this section, we first build a large scale benchmark of taxi origin-destination demand prediction. We then introduce the evaluation metrics of this task and further compare our proposed method with several state-of-the-art methods. Finally, we conduct extensive component analysis to demonstrate the effectiveness of each module of our model.

\subsection{NYC-TOD Dataset}~\label{sec:data}
To the best of our knowledge, there are no public datasets for the citywide taxi origin-destination demand prediction. To evaluate the performances of all compared methods and further promote the relevant research, we also create the first benchmark for this task, denoted as NYC-TOD. It is composed of two data categories, including taxi origin-destination demand data and meteorological data of the New York City in 2014. We choose the data of the last sixty days as the testing set, and all data before that as the training set.

{\bf{Taxi Origin-Destination Demand Data: }}
New York City (NYC) is one of the most prosperous cities in the world and its taxi industry is extremely developed. The origin and destination locations of most NYC taxi trips are in the Manhattan borough~\cite{qian2015characterizing}, therefore we choose the Manhattan as the study area in our work. As discussed in Section~\ref{sec:definition}, we first divide the Manhattan into a $15 \times 5$ grid map based on the longitude and latitude. Each grid represents a geographical region with a size of about $0.75km \times 0.75km$. The detailed partitioned regions of Manhattan are shown in Figure~\ref{fig:NYC_map}.

\begin{figure}
\centering{\includegraphics[width=.40\textwidth]{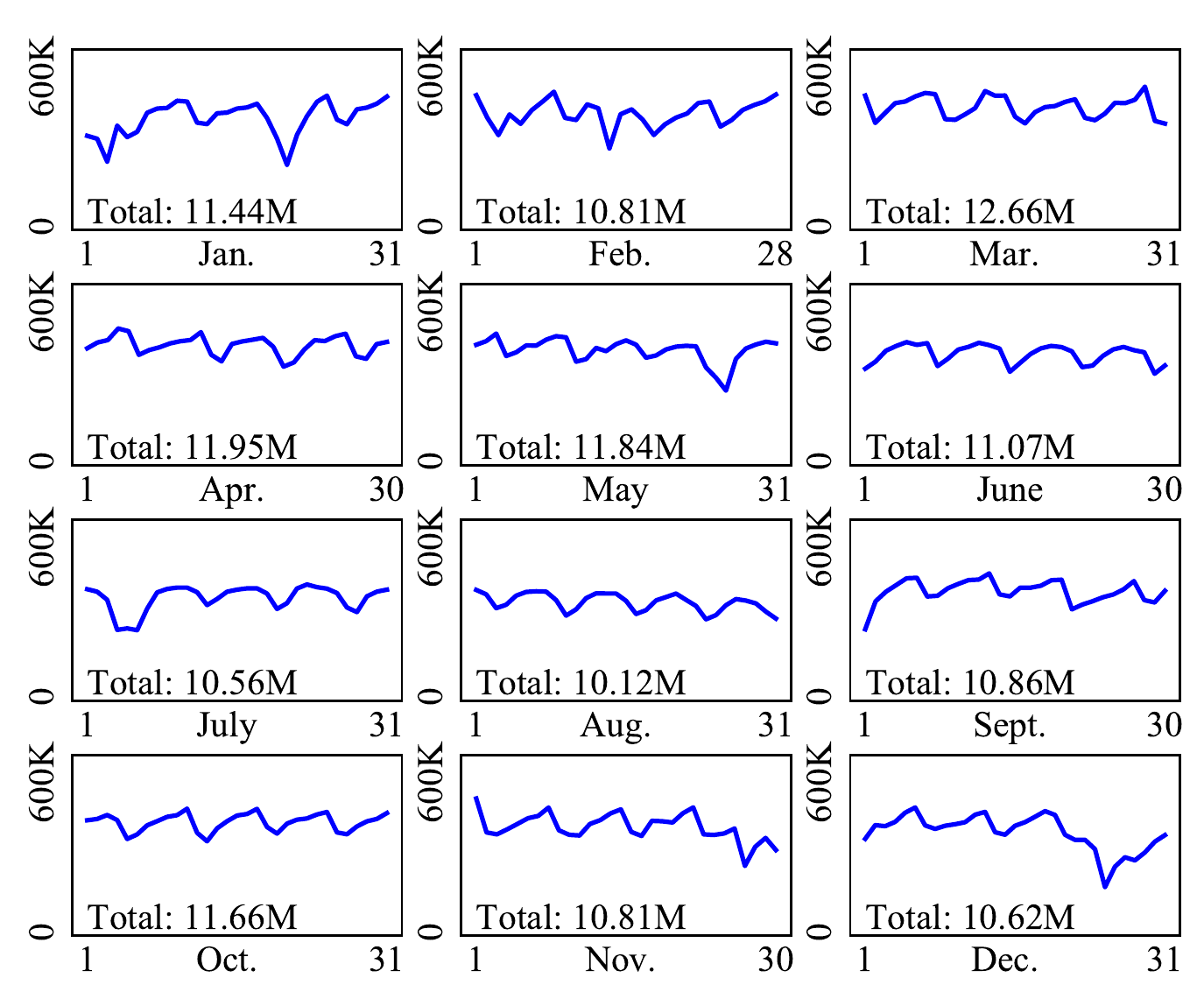}}
\vspace{-3mm}
\caption{The temporal distribution of the taxi demand in our NYC-TOD dataset. We create this dataset with 132 million taxi trip records, more than ten million per month.}
\label{fig:distribution}
\end{figure}

We use the NYC yellow taxi trip records in 2014 to construct our taxi origin-destination demand prediction dataset. These data were collected by the New York City Taxi and Limousine Commission (NYCTLC\footnote{\url{http://www.nyc.gov/html/tlc/html/about/trip_record_data.shtml}}). Each raw trip record contains the timestamp and the geo-coordinates of origin and destination locations. After excluding the trips, of which origin or destination locations aren't in the Manhattan borough, we get 132 million taxi trip records. Finally, we can generate the taxi origin-destination demand matrix in each time interval by calculating the number of taxi trips between all regions according to the timestamps and geo-coordinates of taxi trip records. Each time interval is set to half an hour in this dataset. The total number of taxi demands in each day are summarized in Figure~\ref{fig:distribution}, which shows that more than ten million taxi requests are made in NYC per month. The spatial distribution of taxi demand is shown in Figure~\ref{fig:NYC_map}(b) and we can observe that most taxi demands gather in the city center and traffic hubs.

{\bf{Meteorological Data: }}
We collect the NYC meteorological data in each time interval from Wunderground\footnote{\url{https://www.wunderground.com/}}, which is a well-known meteorological information provider. As the meteorological conditions of all regions are quite similar in the same time interval, we treat the meteorological data observed at the Central Park Station as that of the whole Manhattan borough. We consider the effect of temperature, windchill, humidity, visibility, wind speed, precipitation and weather conditions in our study. The categories of weather condition and the range of other six meteorological indicators are shown in Table~\ref{tab:meteorological_data}. Further, the weather condition is digitized with One-Hot Encoding~\cite{harris2010digital}, while the other six numeric indicators are scaled into the range [0,1] with Min-Max linear normalization. Finally, the meteorological data in time interval ${t}$ can be denoted as a vector ${\bm{M}_t \in R^{29}}$.

\begin{table}[t]
\caption{An Overview of Meteorological Data on NYC-TOD dataset}
\vspace{-2mm}
\newcommand{\tabincell}[2]{\begin{tabular}{@{}#1@{}}#2\end{tabular}}
  \centering
    \begin{tabular}{c|c}
    \hline
    \tabincell{c}{Type} & \tabincell{c}{Information} \\
    \hline
     Temperature / \textcelsius   & [-18.3, 35.6] \\ 
    Windchill / \textcelsius     & [-28.4, 38.5] \\ %
    Humidity / \%                 & [9, 100] \\ %
    Visibility / km               & [0.4, 16.1] \\
    Wind Speed / km/h             & [0.0, 137.0] \\
    Precipitation / mm            & [0.0, 28.7] \\
    Weather Condition           & 23 types(e.g. Sunny, Rainy, Snowy and Unknown) \\
    \hline
    \end{tabular}
\label{tab:meteorological_data}
\end{table}

\subsection{Evaluation Metric}
Following the previous works~\cite{yao2018deep,zhang2017deep}, we adopt the Mean Average Percentage Error (MAPE) and Rooted Mean Square Error (RMSE) as the metrics to evaluate the performance of all methods, which are defined as:
\begin{small}
\begin{equation}
\begin{split}
   MAPE &= \frac{1}{z} \displaystyle\sum_{t=1}^{z} \frac{\|\hat{\bm{X}}_t - \bm{X}_t\|}{\bm{X}_t}, \\
   RMSE &= \sqrt{\frac{1}{z} \displaystyle\sum_{t=1}^{z} \|\hat{\bm{X}}_t - \bm{X}_t\|^2},
\end{split}
\end{equation}
\end{small}%
where ${z}$ is the total number of testing samples, ${\hat{\bm{X}}_t}$ and ${\bm{X}_t}$ are the predicted taxi demand and the corresponding ground truth in time interval ${t}$ respectively. As described in section~\ref{sec:model}, the input and output of our proposed network are normalized into the range ${[-1, 1]}$ during training, so when evaluating, we re-scale the predicted values back to the normal values and then compare them with the ground truth.

In our experiment, we not only evaluate the performance of the task of taxi origin-destination demand prediction, but also consider the task of taxi origin demand prediction. As described in Section~\ref{sec:definition}, the predicted origin demand ${\hat{\bm{O}_t}}$ can be calculated from ${\hat{\bm{X}_t}}$ by ${\sum_{d=0}^{N-1} \hat{\bm{X}_t}(d)}$. For convenience in the following section, the MAPE and RMSE of the former task are denoted as OD-MAPE and OD-RMSE, while these two metrics of the latter task are denoted as O-MAPE and O-RMSE.
When evaluating, we follow the previous work~\cite{yao2018deep} to filter the origin-destination pairs or the origin regions with ground truth less than 5 in each time interval since such low taxi demand is always ignorable in real-world applications.

\subsection{Comparison with the State-of-the-Art}
We compare the performance of our proposed method with the following basic and advanced methods. We tune the parameters of all methods and report their best performance.
\begin{itemize}
  \item \textbf{Historical Average (HA)}: Historical Average predicts the future demand by averaging the historical demands. There are two implemented methods: (1) \textbf{HA-All} averages the historical demands in the same time intervals of every day on the whole training set; (2) \textbf{HA-Rec} averages the taxi demands of previous ${n}$ time intervals.
  \item \textbf{Linear Regression}: We implement two typical linear regression methods: Ordinary Least Squares Regression (\textbf{OLSR}~\cite{craven2011ordinary}) and \textbf{Lasso} Regression~\cite{tibshirani1996regression} with $\ell_{1}$-norm regularization. They take the concatenation of the demand matrices of previous ${n}$ time intervals as input and predict the taxi demand between any two regions.
  \item \textbf{XGBoost~\cite{chen2016xgboost}}: XGBoost is a powerful boosting trees based method. Similar with OLSR and Lasso, XGBoost concatenates the demand matrices of previous ${n}$ time intervals and takes them to forecast the taxi OD demand.
  \item \textbf{Multiple Layer Perceptron (MLP)}: A neural network consists of four fully connected layers with 128, 128, 64 and 75 neurons respectively. The MLP forecasts the every channel of $X_t$ by taking the corresponding channels of demand matrices of previous ${n}$ time intervals as input.
  \item \textbf{ST-ResNet~\cite{zhang2017deep}}: ST-ResNet is a deep learning based method that predicts the future traffic inflow and outflow. We utilize its released code\footnote{\url{https://github.com/lucktroy/DeepST/tree/master/scripts/papers/AAAI17}} to predict the taxi origin-destination demand.
  \item \textbf{ConvLSTM~\cite{xingjian2015convolutional}}: ConvLSTM is our LSC module + TEC module. Specifically, the LSC module in this network only contains the origin view ConvNet and takes $X_i$ as input to learn the local spatial context.
\end{itemize}

\begin{table}
  \caption{PERFORMANCE OF DIFFERENT METHODS \protect\\ \bf{ON THE WHOLE NYC-TOD TESTING SET}}
  \vspace{-2mm}
  \label{tab:NYC-Performance}
  \centering
  \begin{tabular}{ccccc}
    \toprule
    Method & OD-MAPE & OD-RMSE & O-MAPE & O-RMSE  \\
    \midrule
    HA-All   & 37.71\% & 1.93 & 45.04\% & 52.44\\
    HA-Rec & 35.46\% & 1.89 & 47.59\% & 54.33\\
    Lasso        & 33.85\% & 1.65 & 34.89\% & 33.00\\
    OLSR           & 33.86\% & 1.65 & 33.09\% & 32.68\\
    XGBoost      & 32.04\% & 1.54 & 37.78\% & 31.23\\
    MLP          & 30.70\% & 1.49 & 25.24\% & 25.60\\
    ST-ResNet    & 28.53\% & 1.38 & 24.16\% & 22.43\\
    ConvLSTM     & 27.99\% & 1.36 & 19.89\% & 21.02\\
    CSTN         & {\bf\textcolor{red}{27.37\%}} & {\bf\textcolor{red}{1.32}} & {\bf\textcolor{red}{18.48\%}} & {\bf\textcolor{red}{19.85}}\\
    \bottomrule
\end{tabular}
\end{table}

\begin{figure*}[ht]
  \centerline{
    \includegraphics[width=1\columnwidth,height=5.2cm]{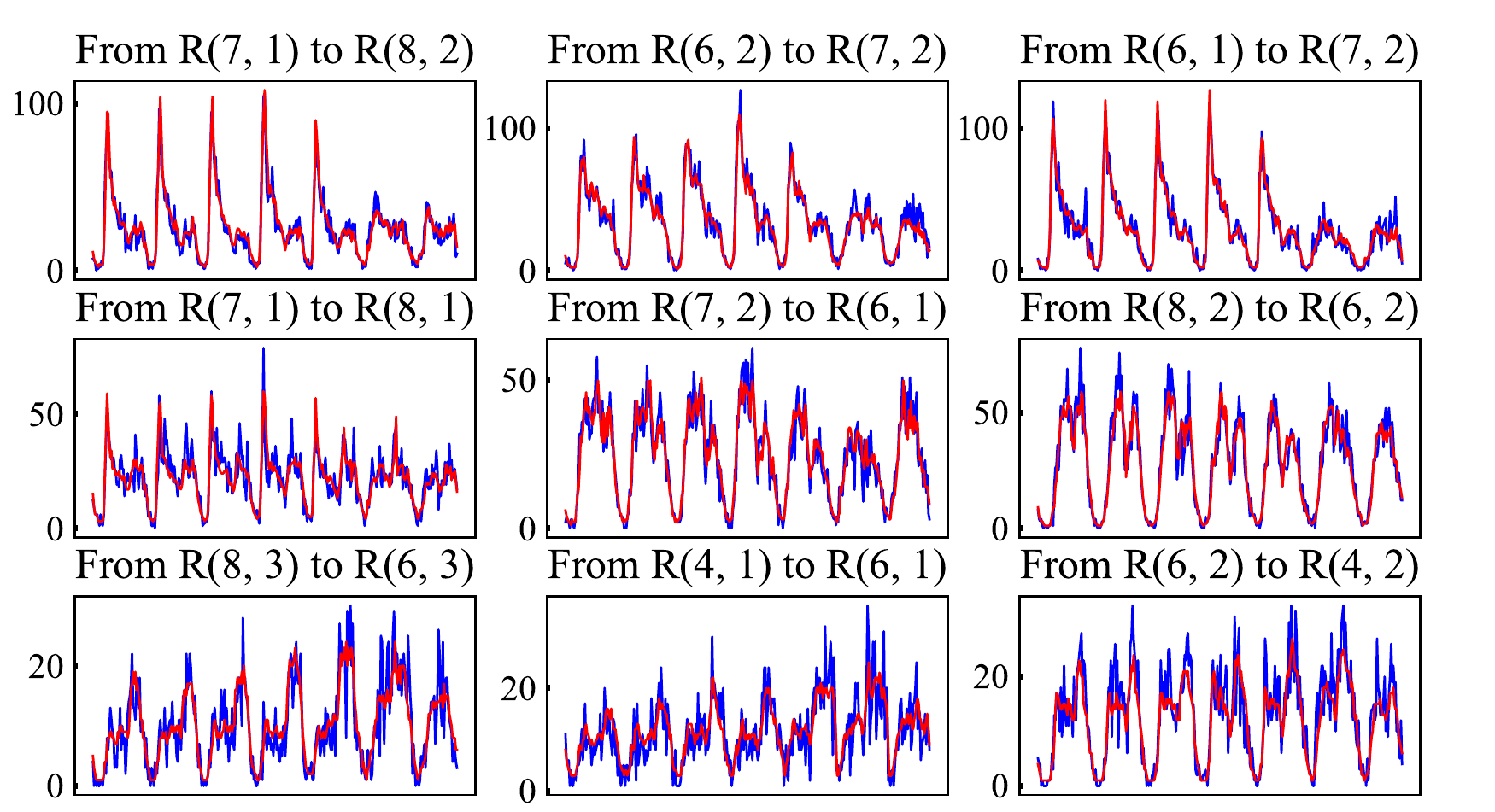}
    \includegraphics[width=1\columnwidth,height=5.2cm]{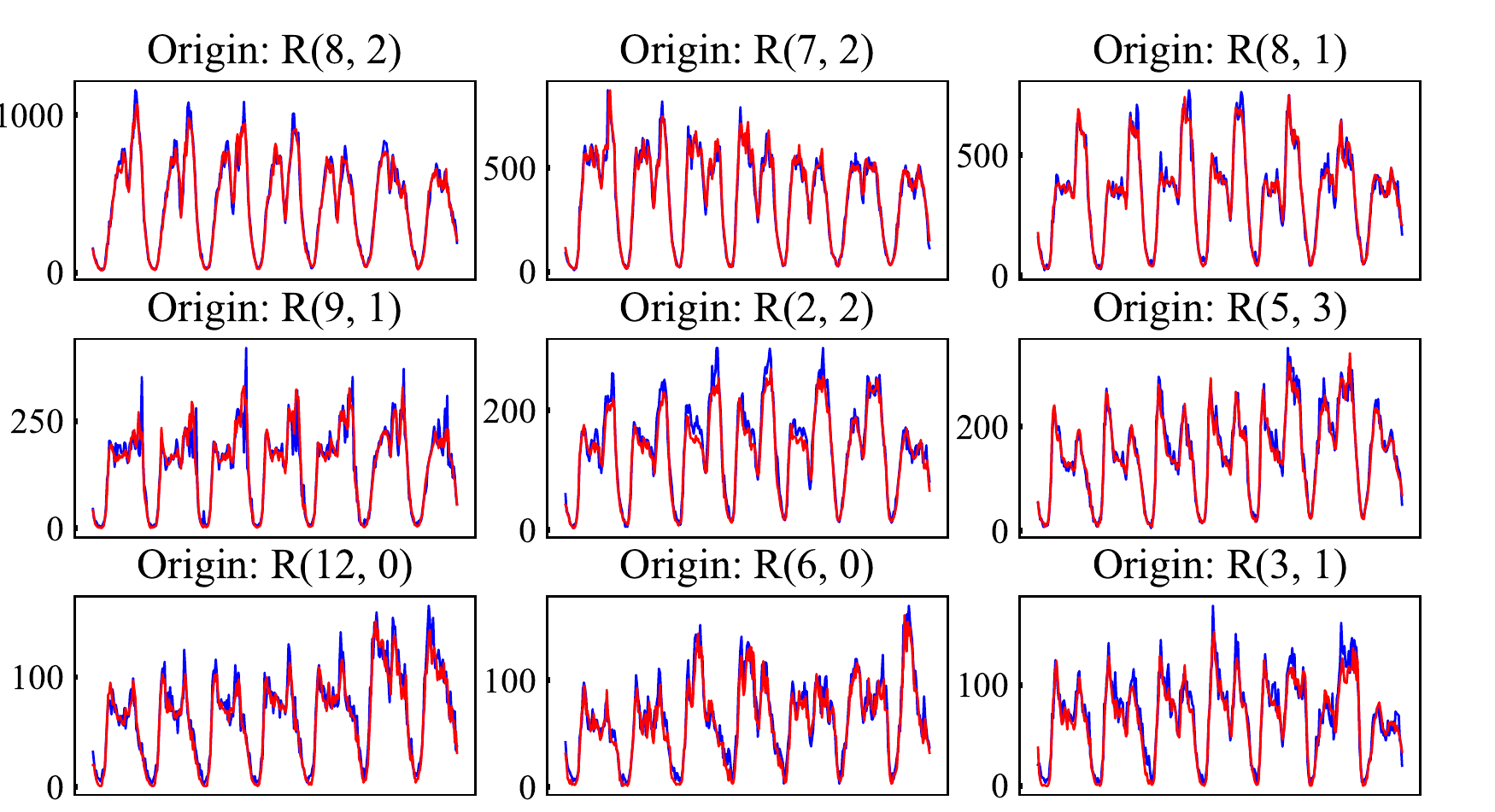}
  }
  \vspace{-2 mm}
  \caption{Visualization of the estimated taxi interregional demands (left) and origin demands (right) within one week (Nov 3-9, 2014).
  For each subfigure, the top row is the result of regions with high demands, while the bottom two rows are the result of regions with middle and low demands respectively. The red lines are {\color{red}our predicted result} and the blue lines are {\color{blue}the ground truth}.}
\label{fig:taxi_predicted_result}
\end{figure*}

\textbf{Performance on the Whole Testing Set: }
We first conduct the comparison of our proposed method with other methods on the whole NYC-TOD testing set. The results of all methods are summarized in Table~\ref{tab:NYC-Performance} and it can be observed that our method outperforms other competed methods by a margin. Specifically, our method achieves the lowest MAPE and RMSE on the task of taxi origin-destination demand prediction. Moreover, for the taxi origin demand prediction, our method achieves 7.1\% and 5.6\% relative performance improvements over O-MAPE and O-RMSE, compared to the existing best-performing method ConvLSTM. Figure~\ref{fig:taxi_predicted_result} shows the taxi origin-destination demands and taxi origin demands predicted by our CSTN. We can observe that our method is robust to forecast the taxi demands of different scale.

Despite some competed methods (such as MLP, ST-ResNet and ConvLSTM) also adopt deep learning techniques to predict the taxi demand, they perform worse than our CSTN. The main reasons are that MLP fails to capture the local spatial context and ST-ResNet does not explicitly learn temporal evolution context, while ConvLSTM does not model the global correlation context. Compared with these methods, our method integrates the above various context into a unified framework to predict the taxi demand in future time intervals.

\begin{table}
  \caption{PERFORMANCE OF DIFFERENT METHODS \protect\\ \bf{ON THE HIGH-DEMAND REGIONS}}
  \vspace{-2mm}
  \label{tab:high-demand-regions}
  \centering
  \begin{tabular}{ccccc}
    \toprule
    Method & OD-MAPE & OD-RMSE & O-MAPE & O-RMSE  \\
    \midrule
    HA-All   & 36.96\% & 5.69 & 46.47\% & 93.38\\
    HA-Rec   & 35.65\% & 5.67 & 49.62\% & 97.16\\
    Lasso        & 31.51\% & 4.59 & 24.88\% & 57.32\\
    OLSR         & 31.55\% & 4.58 & 24.28\% & 56.80\\
    XGBoost      & 29.63\% & 4.28 & 34.30\% & 53.20\\
    MLP          & 27.81\% & 4.01 & 17.18\% & 42.15\\
    ST-ResNet    & 25.98\% & 3.71 & 16.13\% & 37.09\\
    ConvLSTM     & 25.81\% & 3.65  & 13.80\% & 35.33\\
    CSTN         & {\bf\textcolor{red}{24.93\%}} & {\bf\textcolor{red}{3.58}} & {\bf\textcolor{red}{12.92\%}} & {\bf\textcolor{red}{33.73}}\\
    \bottomrule
\end{tabular}
\end{table}

\textbf{Performance on the High-Demand Regions: }
As shown in Figure~\ref{fig:NYC_map}(b), the spatial distribution of taxi demand is not uniform and most of the taxi demands are gathered in some regions, therefore the taxicab companies may give priority to meet the taxi demand of these regions. In this section, we evaluate the performance of all compared methods on the high-demand regions. We first measure the taxi origin demand of each region on the whole training set of NYC-TOD and then choose twenty regions with the highest demands. These regions cover about 70\% of the taxi demand in Manhattan.
We only evaluate the origin-destination demand between these regions and the origin demand within these regions. As shown in Table~\ref{tab:high-demand-regions}, our method achieves the best performance in comparison to other methods on high-demand regions. Specifically, our method outperforms ConvLSTM by about 1\% over the MAPE metric for two types of taxi demand prediction.

\begin{figure}
\includegraphics[width=0.50\textwidth]{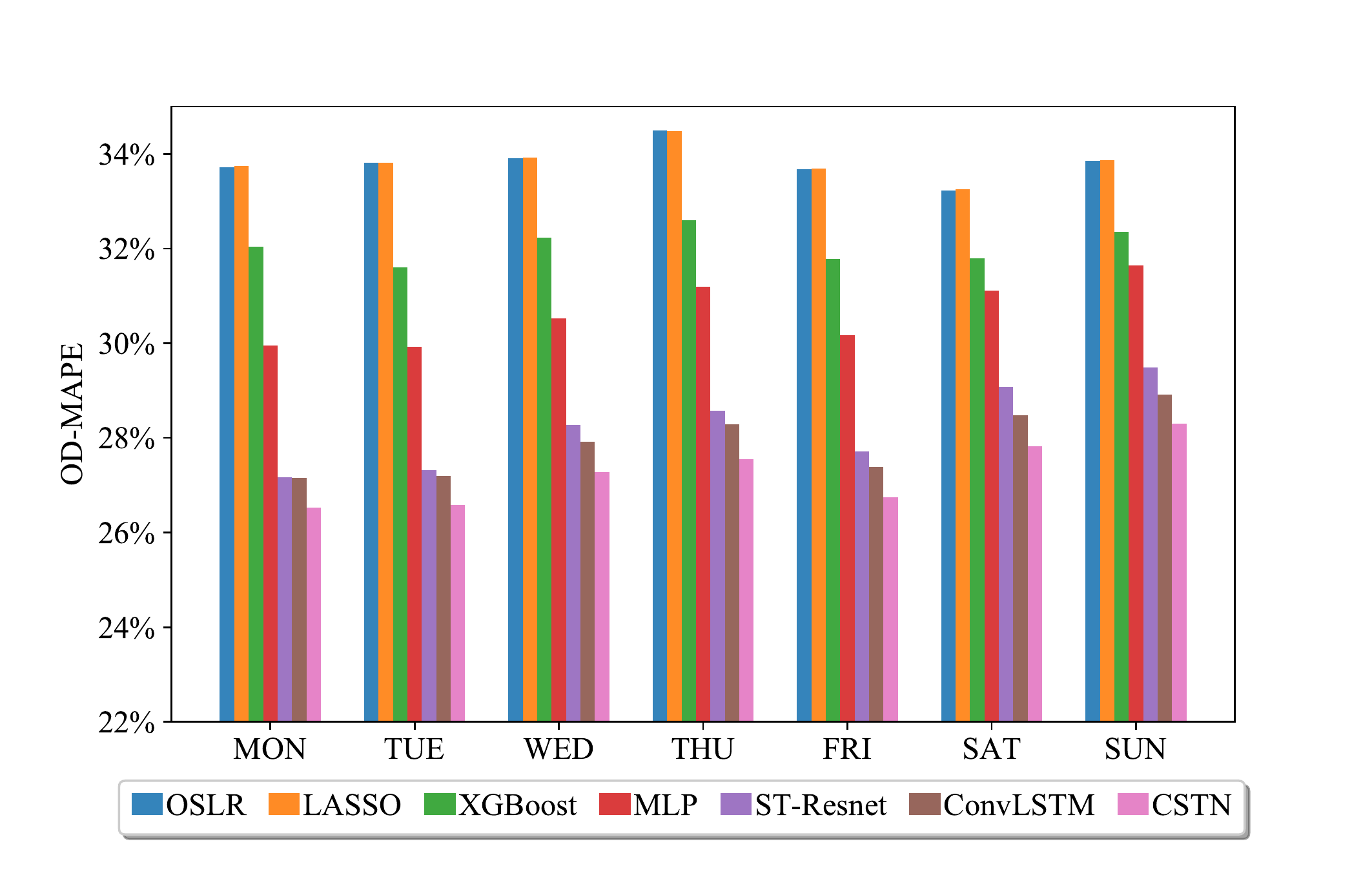}
\vspace{-6mm}
\caption{The MAPE of different methods for taxi origin-destination demand prediction on different days of the week. Our method consistently outperforms the other methods in all days of the week.}
\label{fig:week_diff}
\end{figure}

\begin{table}
  \caption{The MAPE of different methods for taxi origin-destination demand prediction on weekdays and weekends}
  \vspace{-2mm}
  \label{tab:weekdays-weekends}
  \centering
  \begin{tabular}{ccccc}
    \toprule
    Method & Weekdays & Weekends \\
    \midrule
    Lasso        & 33.93\% & 33.56\% \\
    OLSR           & 33.92\% & 33.54\% \\
    XGBoost      & 32.05\% & 32.07\% \\
    MLP          & 30.35\% & 31.38\% \\
    ST-ResNet    & 29.44\% & 31.34\% \\
    ConvLSTM     & 27.58\% & 28.70\% \\
    CSTN         & {\bf\textcolor{red}{26.93\%}} & {\bf\textcolor{red}{28.06\%}} \\
    \bottomrule
\end{tabular}
\end{table}

\textbf{Performance on Different Days}
We compare the performance of all methods on different days of the week in this section. We will exclude the result of HA-All and HA-Rec in the following experiment as they are of very poor performance. Here we only report their performance over the OD-MAPE metric, and the similar phenomenon also occurs over the O-MAPE.
As shown in Figure~\ref{fig:week_diff}, our method consistently outperforms other competed methods in all days of the week. Furthermore, we also average the performance of all methods on weekdays and weekends. The results are summarized in Table~\ref{tab:weekdays-weekends} and our method still achieves the best performance.
We can observe that the performances of three shallow methods on weekdays and weekends are comparable. In contrast, the performance of four deep learning based methods on weekdays is better than that on weekends. Yao et al.\cite{yao2018deep} also found this phenomenon and one main reason is that the taxi demand patterns are less regular on weekends. We can conclude that the deep learning based methods have more capacity to capture the regular patterns on weekdays while learning the inconspicuous patterns on weekends.

\subsection{Component Analysis}

\begin{table}
  \centering
  \caption{Comparison of Taxi Demand Prediction with Different Context}
  \vspace{-2mm}
  \label{tab:different_context}
  \begin{tabular}{ccc}
    \toprule
    Method & OD-MAPE & O-MAPE \\
    \midrule
    LSC Net         & 28.54\% & 20.80\% \\
    LSC+TEC  Net     & 27.80\% & 19.41\% \\
    LSC+TEC+GCC  Net & 27.27\% & 18.48\% \\
    \bottomrule
\end{tabular}
\end{table}

\textbf{Influence of Different Context: }
Our full model consists of three components for three types of context modeling. To explore the influence of different context on taxi demand prediction, we implement the following variants of our model with different components:
\begin{itemize}
  \item \textbf{LSC Net}: This network only contains the LSC module and it directly concatenates the local spatial features of each time interval to predict the future taxi demand with a convolutional layer.
  \item \textbf{LSC+TEC Net}: This network contains the LSC module and TEC module, but without the GCC module. It feeds the last hidden state of the TEC module into a convolutional layer to predict the taxi demand.
  \item \textbf{LSC+TEC+GCC Net}: As the full version of CSTN, this network integrates the local spatial context, temporal evolution context and global correlation context to predict the taxi demand.
\end{itemize}

As shown in Table~\ref{tab:different_context}, the LSC Net achieves an OD-MAPE of 28.54\% and an O-MAPE of 20.80\%. It outperforms the ST-ResNet which has more convolutional layers, as our LSC module adequately captures the local spatial context with the Two-View ConvNet.
When explicitly modeling the temporal evolution context of taxi demand with LSTM, the LSC+TEC Net gets an OD-MAPE of 27.80\% and an O-MAPE of 19.41\%, achieving an obvious performance improvement compared to the LSC Net.
After integrating the global correlation context with the GCC module, the LSC+TEC+GCC Net can further decrease the OD-MAPE to 27.27\% and OD-MAPE to 18.48\%, with 2.5\% relative performance improvement on average.
The experimental result shows that our network can achieve notable performance improvement by modeling these context, which also indicates the effectiveness of these context for the task of taxi demand prediction.

\begin{table}
  \centering
  \caption{Effectiveness of the Two-View ConvNet in the LSC module}
  \vspace{-2mm}
  \label{tab:LSC-variant}
  \begin{tabular}{ccc}
    \toprule
    Method & OD-MAPE & O-MAPE  \\
    \midrule
    Origin View      & 28.94\% & 23.03\% \\
    Origin View + Destination view       & 28.54\% & 20.80\% \\
    \bottomrule
\end{tabular}
\end{table}

\textbf{Effectiveness of the Two-View ConvNet in the LSC module: }
As described in Section~\ref{sec:LSC}, we use a Two-View ConvNet to model the local spatial context from origin view and destination view. To validate the effectiveness of the Two-View ConvNet, we train a variant of LSC Net that only takes OD matrix $\bm{X}_i$ as input to learn the local spatial context from origin view.
As shown in Table~\ref{tab:LSC-variant}, only with origin view ConvNet, the LSC Net performs so poorly. After adding the destination view, the performance will be improved with 0.5\% and 2.03\% over metrics OD-MAPE and O-MAPE respectively. The experimental result shows that the destination view context is also beneficial for taxi demand and our LSC module can capture the local spatial context effectively.

\begin{table}
  \centering
  \caption{Influence of Local and Global Context for Taxi Demand Prediction}
  \vspace{-2mm}
  \label{tab:Local_Global_Fusion}
  \begin{tabular}{ccc}
    \toprule
    Method & OD-MAPE & O-MAPE  \\
    \midrule
    Global       & 28.56\% & 20.25\% \\
    Local        & 27.80\% & 19.41\% \\
    Local+Global & 27.37\% & 18.48\% \\
    \bottomrule
\end{tabular}
\end{table}

\textbf{Influence of Local and Global Context: }
As described in Section~\ref{sec:GCC}, we forecast the taxi demand with the concatenation of local feature ${F^{lt}}$ and global feature ${F^g}$. To analyze how these two features contribute to the performance, we train other two variants of CSTN to predict the taxi demand only with ${\bm{F}^{lt}}$ or ${\bm{F}^g}$.
As shown in Table~\ref{tab:Local_Global_Fusion}, the performance of the local feature ${\bm{F}^{lt}}$ is better than that of the global feature ${\bm{F}^g}$, which indicates the local feature is more efficient for this task.
When combining the local and global feature for the final prediction, our method achieves the best performance, which shows that the local context and global context are complementary for the taxi demand prediction.

\textbf{Influence of Meteorology: }
In this section, we will explore the influence of meteorology on taxi demand prediction. We train another LSC+TEC Net and CSTN without considering the meteorology.
As shown in table~\ref{tab:meteorology}, without taking the meteorological data into consideration, the LSC+TEC Net and CSTN respectively get an O-MAPE of 20.03\% and 19.72\%. In contrast, when predicting the taxi demand with meteorological data, the LSC+TEC Net and CSTN can decrease the O-MAPE to 19.41\% and 18.48\%, with 3.1\% and 6.23\% relative performance improvement.
Moreover, we also verify the relevance of each variable in meteorological data for the taxi demand prediction by filtering it when inferring. When exploring the effect of weather condition, we set it to the type \textit{Unknown} for all time intervals. For each of the other numeric indicators, it is set to its mean value of the whole training set. The changes in performance are shown in table~\ref{tab:meteorological_variables_relevance} and we can see that the OD-MAPE and O-MAPE are incremental to some extent when filtering the meteorological data variables.
These experiments show that the meteorological information can help to improve the performance of taxi demand prediction.

\begin{table}
  \caption{Comparison of the Taxi Demand Prediction with or without Meteorology}
  \vspace{-2mm}
  \centering
  \begin{tabular}{ccc}
    \toprule
    Method & OD-MAPE & O-MAPE  \\
    \midrule
    LSC+TEC Net W/O Meteorology         & 28.08\% & 20.03\% \\
    LSC+TEC Net W/- Meteorology          & 27.80\% & 19.41\% \\
    \midrule
    CSTN W/O Meteorology        & 27.69\% & 19.72\% \\
    CSTN W/- Meteorology         & 27.37\% & 18.48\% \\
    \bottomrule
  \end{tabular}
  \label{tab:meteorology}
\end{table}

\begin{table}
  \caption{The Incremental Error \protect\\ when filtering each variable in meteorological data}
  \vspace{-2mm}
  \centering
  \begin{tabular}{ccc}
    \toprule
    Variables & OD-MAPE & O-MAPE  \\
    \midrule
    Temperature         & 0.054\% & 0.352\% \\
    Windchill           & 0.056\% & 0.354\% \\
    Humidity            & 0.042\% & 0.315\% \\
    Visibility          & 0.030\% & 0.244\% \\
    Wind Speed          & 0.032\% & 0.246\% \\
    Precipitation       & 0.033\% & 0.259\% \\
    Weather Condition   & 0.032\% & 0.069\%  \\
    \bottomrule
  \end{tabular}
  \label{tab:meteorological_variables_relevance}
\end{table}

\textbf{Influence of Sequence Length: }
As described in Section~\ref{sec:model}, we can implement our model with different sequence length ${n}$ of time intervals. To explore the influence of sequence length, we train our model with different ${n}$.
As shown in table~\ref{tab:seqlen}, the OD-MAPE and O-MAPE gradually decrease as the sequence length increases. Our method achieves the best performance with five time intervals (2.5 hours) and longer sequence hardly results in obvious performance improvement.
One potential reason is that the future taxi demand is more relevant to the short-term tendency. Therefore, feeding too long time sequence into the network no longer helps to boost the performance and we finally set the sequence length ${n}$ to 5 in all experiments.

\begin{table}
  \caption{Comparison of the Taxi Demand Prediction with Different Sequence Length of the Time Intervals}
  \vspace{-2mm}
  \centering
  \begin{tabular}{ccc}
    \toprule
    Length & OD-MAPE & O-MAPE  \\
    \midrule
    2          & 30.06\% & 28.86\% \\
    3          & 29.02\% & 25.62\% \\
    4          & 28.04\% & 22.84\% \\
    5          & 27.37\% & 18.48\% \\
    6          & 27.61\% & 19.10\% \\
    \bottomrule
  \end{tabular}
  \label{tab:seqlen}
\end{table}

\subsection{Further Discussion}

\begin{table}
  \caption{Comparison of Running Times of Deep Models \protect\\ on an NVIDIA 1080 GPU}
  \vspace{-2mm}
  \label{tab:time}
  \centering
  \begin{tabular}{cc}
    \toprule
    Method & Time/ms \\
    \midrule
    MLP          & 0.329  \\
    ST-ResNet    & 0.399  \\
    ConvLSTM     & 0.739  \\
    CSTN         & 1.187  \\
    \bottomrule
\end{tabular}
\end{table}

\textbf{Runtime Efficiency: }
In this subsection, we compare the running times of different methods for taxi origin-destination demand prediction.
As shown in Table~\ref{tab:time}, all deep learning based methods can achieve practical runtime efficiencies on an NVIDIA 1080 GPU. Specifically, our CSTN only costs 1.187 ms to predict the taxi demand of next time interval, which is totally acceptable in the industrial community. As for the traditional methods (such as Lasso, OLSR and XGBoost) of CPU implementation, we evaluate their running times on Intel Xeon 2.40GHz E5-2620 CPU. Lasso and OLSR can conduct a prediction within 0.381 ms, while XGBoost requires 5.666 ms for each inference, as it processes the prediction of each region pair independently.
In summary, all compared methods can perform in real-time and the runtime efficiency is not the bottleneck of this task, thus we should pay more attention to improving the accuracy of the taxi demand prediction.

\begin{figure}[t]
\centerline{\includegraphics[width=0.45\textwidth]{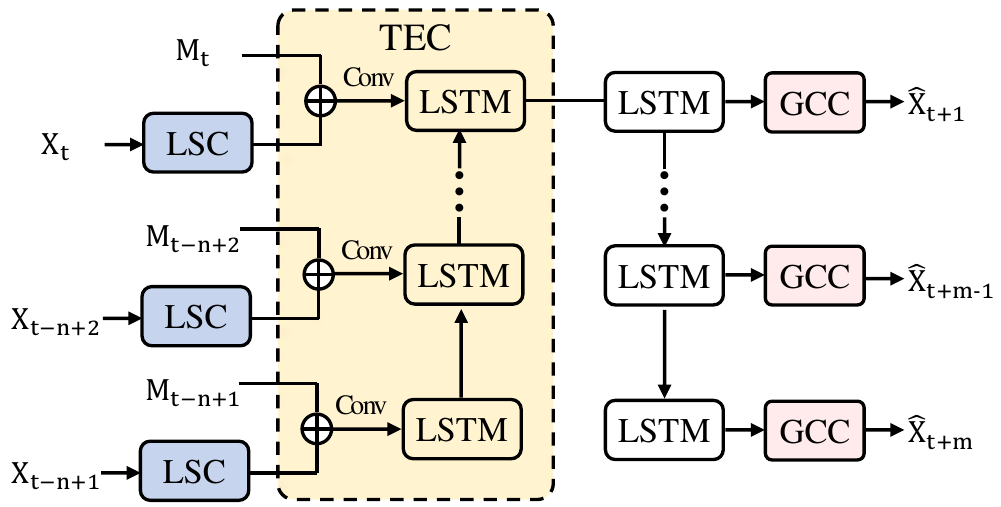}}
\vspace{-2mm}
\caption{The architecture of the L-CSTN for long-term taxi demand prediction. Compared with the original CSTN, L-CSTN forecasts the taxi demand of the multiple time intervals in the future with extra $m$ decoding ConvLSTM units.}
\label{fig:long_term_prediction}
\end{figure}

\textbf{Long-Term Taxi Demand Prediction: }
In this subsection, we extend our CSTN to predict the long-term taxi demand. Here, we take the historical demand data $\{ \bm{X}_i | i = t-n,...,t\}$ and meteorological data $\{ \bm{M}_i | i = t-n+1,...,t\}$ to forecast the future demand $\{\hat{\bm{X}}_i | i = t+1,...,t+m\}$, where $n$ and $m$ are set to 5 and 6 respectively in our experiment. The long-term prediction version of CSTN is denoted as L-CSTN and its architecture is shown in Figure~\ref{fig:long_term_prediction}. L-CSTN first encodes the historical data with LSC module and generates the feature $\bm{F}^{lt}$ with TEC module. Then, $\bm{F}^{lt}$ is fed into $m$ decoding ConvLSTM units and each of them is followed by a GCC module to predict the taxi demand. The performance of long-term taxi demand prediction is shown in Table~\ref{tab:long_term_prediction}. Our L-CSTN achieves an OD-MAPE of 28.63\% and an O-MAPE of 20.50\% for the demand $\hat{\bm{X}}_{t+2}$. As the predicted time intervals increase, the performance gradually drops. For the demand $\hat{\bm{X}}_{t+6}$, despite its OD-MAPE and D-MAPE increase to 30.86\% and 24.85\%, this estimated result is still very practical for taxi preallocation.

\begin{table}
  \caption{Performance of the Long-Term Taxi Demand Prediction}
  \vspace{-2mm}
  \centering
  \begin{tabular}{cccc}
    \toprule
    Model & Demand & OD-MAPE & O-MAPE  \\
    \midrule
    CSTN                    & $\hat{X}_{t+1}$          & 27.37\% & 18.48\% \\
    \hline
    \multirow{5}{*}{L-CSTN}
    & $\hat{X}_{t+2}$      & 28.63\% & 20.50\% \\
    & $\hat{X}_{t+3}$      & 29.14\% & 21.83\% \\
    & $\hat{X}_{t+4}$      & 29.75\% & 22.79\% \\
    & $\hat{X}_{t+5}$      & 30.44\% & 24.78\% \\
    & $\hat{X}_{t+6}$      & 30.86\% & 24.85\% \\
    \bottomrule
  \end{tabular}
  \label{tab:long_term_prediction}
\end{table}

\begin{figure}
\centerline{\includegraphics[width=.420\textwidth]{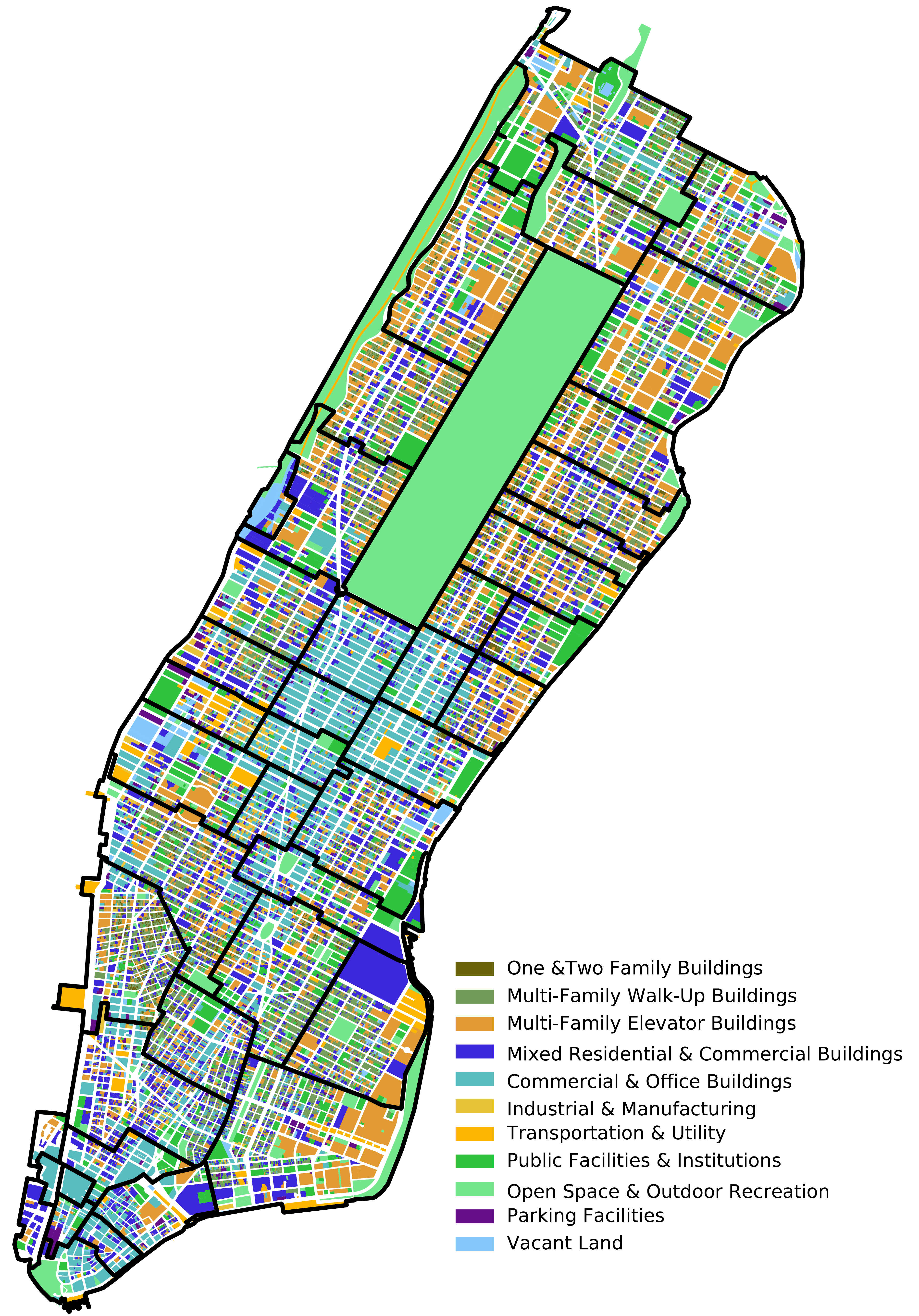}}
\vspace{-2mm}
\caption{Illustration of the region partition of NYC on the basis of Zip Code Tabular (ZCT) and land use homogeneity.}
\label{fig:land_use_homogeneity_partition}
\end{figure}

\textbf{Different Region Partition Manners: } In this subsection, we explore the performance of different region partition manners. Geographical coordinate (longitude and latitude) is widely used to generate rectangular regions~\cite{zhang2017deep,yao2018deep,liu2018attentive}, while land use homogeneity is another good foundation of region partition. According to the Pluto (Primary Land Use Tax Lot Output dataset\footnote{\url{https://www1.nyc.gov/site/planning/data-maps/open-data/dwn-pluto-mappluto.page}}), we visualize the land types of each building block of the Manhattan borough in Figure~\ref{fig:land_use_homogeneity_partition} and find there may exist multiple categories of land in a local region. Inspired by the previous work~\cite{qian2017forecasting}, we first generate multiple areas on the basis of the Zip Code Tabular (ZCT) and then manually adjust their spaces with the land use homogeneity. The final 44 regions with different shapes are shown in Figure~\ref{fig:land_use_homogeneity_partition} and each region has relatively consistent land use homogeneity. In this case, the historical taxi origin-destination demand $\bm{X}_t$ in time
interval $t$ is organized as a 2D matrix with a dimension of $44 \times 44$ and $\bm{X}_t(i,j)$ denotes the demand from origin
region $i$ to destination region $j$. We reconstruct the NYC-TOD Dataset with the new region coordinates and retrain the compared methods. Specifically, since $\bm{X}_t$ lacks the spatial information, our CSTN utilizes a CNN with four convolutional layers to encode the $\bm{X}_t$ and then feed the feature to the TEC module.

The OD-MAPE of four deep learning based methods of different region partition manners is shown in Table~\ref{tab:different_partition}. We can observe that convincing performance can be achieved by the manner ``ZCT + Land Use Homogeneity", but it is still slightly worse than the manner ``Geographical Coordinate". The main reason is that the data organization format $\bm{X}_t \in R^{N \times N}$ of ``ZCT + Land Use Homogeneity" can not well preserve the local spatial information of the taxi demand, where $N$ is the total number of regions.
How to boost the performance with the local spatial information and land use homogeneity is worth exploring in the future works.

\begin{table}
  \caption{The OD-MAPE of Different Region Partition Manners}
  \vspace{-2mm}
  \label{tab:different_partition}
  \centering
  \begin{tabular}{c|c|c}
    \hline
    \multirow{2}{*}{Manner} & \multirow{2}{*}{Geographical Coordinate} & ZCT + \\
     & & Land Use Homogeneity \\
    \hline
    MLP          & 30.70\% & 30.78\% \\
    ST-ResNet    & 28.53\% & 28.82\% \\
    ConvLSTM     & 27.99\% & 28.54\% \\
    CSTN         & 27.37\% & 27.91\% \\
    \hline
  \end{tabular}
\end{table}

\section{Conclusion}~\label{sec:conclusion}
In this paper, we introduce a more worth-exploring task, taxi origin-destination demand prediction, which aims at predicting the taxi demand between all regions in the future time intervals. We argue that the information of passengers¡¯ destinations is also critical for the taxi preallocation systems, since some factors (e.g. the city management rules and the individual preference of drivers) may affect the supply amount of available taxi between two regions as mentioned in Section~\ref{sec:introduction}. Therefore, it's essential to combine the predicted taxi OD demand and the aforementioned external factors to optimize the taxi preallocation scheme.

We address this problem with a Contextualized Spatial-Temporal Network (CSTN), which integrates local spatial context, temporal evolution context and global correlation context in one united framework. By learning the taxi demand patterns from historical data, the proposed CSTN can make taxi demand predictions for all regions pairs. 132 million taxi trip records of New York City is used to train and evaluate our model. Experimental results show that our model achieves an OD-MAPE of 24.93\% and an O-MAPE of 12.92\%, outperforming other state-of-the-art methods on both tasks of taxi OD demand prediction and origin demand prediction.
Further, we extend our CSTN to predict the long-term taxi demand and our method achieves very practical performance.

How to divide a city into different regions is still an open problem. In the future work, we will explore a better region partition manner, with which the spatial information and land use homogeneity information can be efficiently used simultaneously. Meanwhile, our work can be extended by adding more information to the network, such as the periodic taxi demand and the Point of Interest (POI) in each region, which may help to further boost the performance. Finally, we will cooperate with some taxicab requesting platforms and optimize their taxi preallocation systems with the prediction OD demand and the aforementioned external factors. Such systems are expected to decrease the inefficient operations of the taxi industry.


\ifCLASSOPTIONcaptionsoff
  \newpage
\fi

\bibliographystyle{IEEEtran}
\bibliography{OD_demand}


\begin{IEEEbiography}[{\includegraphics[width=1in,height=1.25in,clip,keepaspectratio]{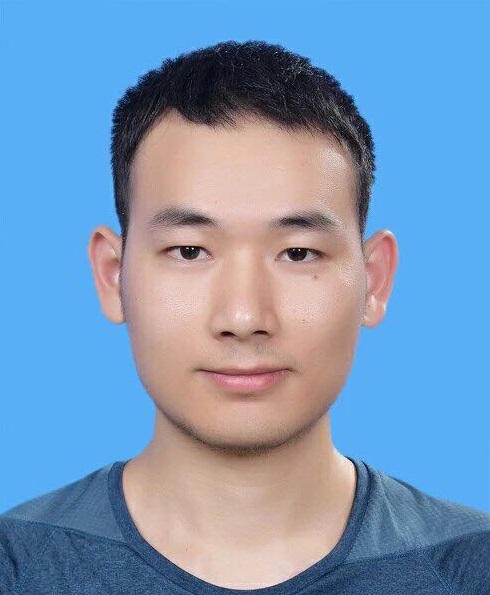}}]
{Lingbo Liu} received the B.E. degree from the School of Software, Sun Yat-sen University, Guangzhou, China, in 2015, where he is currently pursuing the Ph.D degree in computer science with the School of Data and Computer Science. His current research interests include computer vision, intelligent transportation systems, and urban computing.
\end{IEEEbiography}

\begin{IEEEbiography}[{\includegraphics[width=1in,height=1.25in,clip,keepaspectratio]{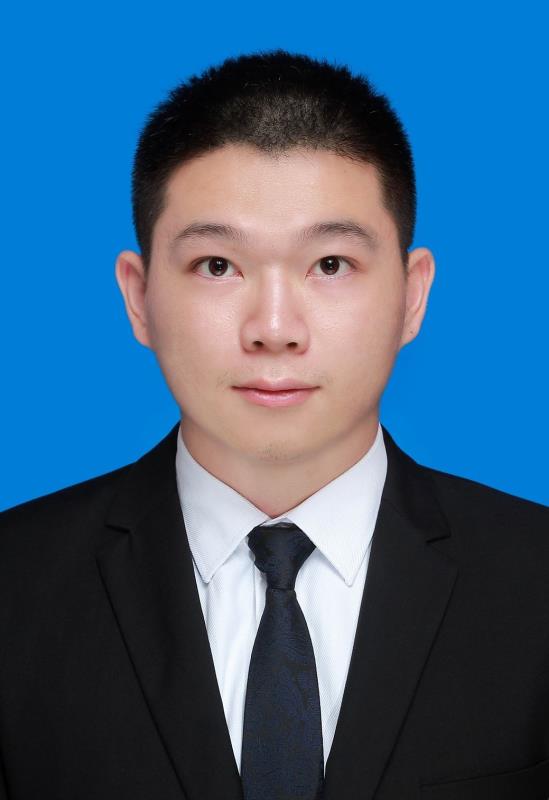}}]
{Zhilin Qiu} received the B.E. degree from the School of Software, Sun Yat-sen University, Guangzhou, China, in 2016, where he is currently pursuing the Master's degree in computer science with the School of Data and Computer Science.
His current research interests include computer vision, intelligent transportation systems, and parallel computation.
\end{IEEEbiography}

\begin{IEEEbiography}[{\includegraphics[width=1in,height=1.25in,clip,keepaspectratio]{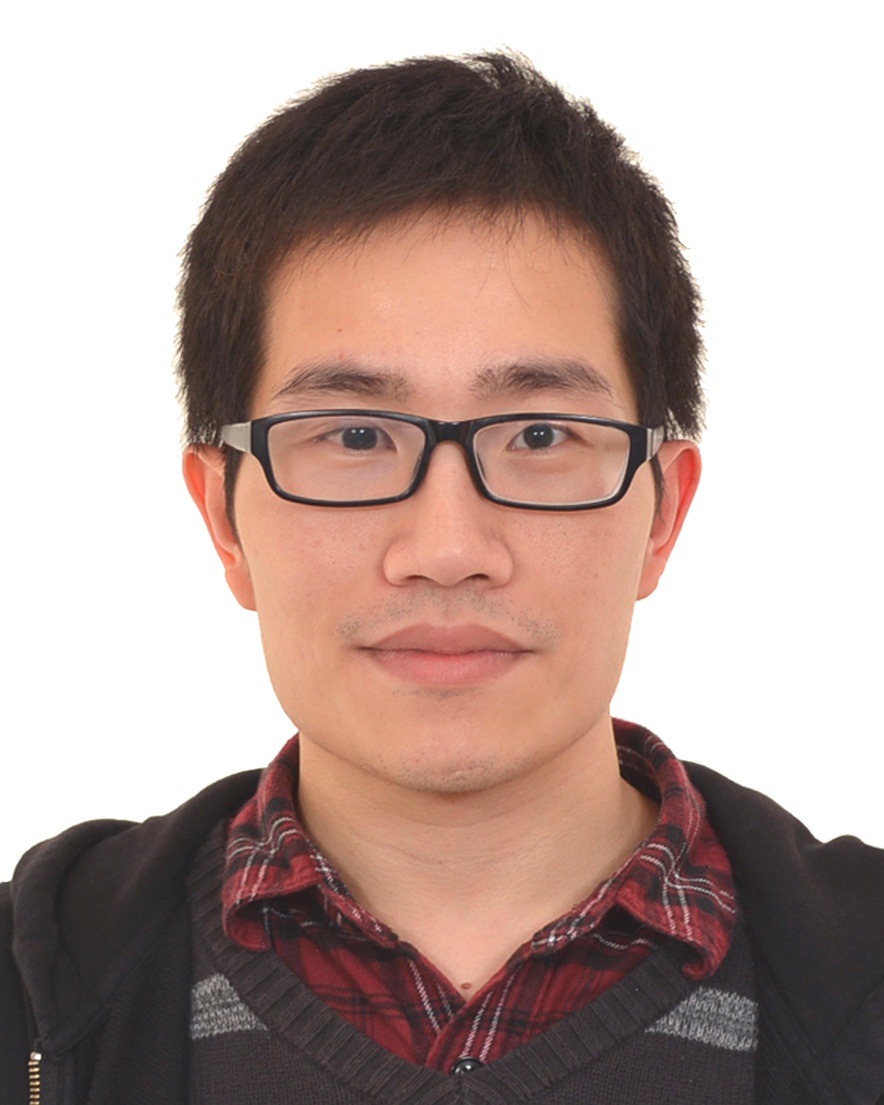}}]
{Guanbin Li} is currently a research associate professor in School of Data and Computer Science, Sun Yat-sen University. He received his PhD degree from the University of Hong Kong in 2016. He was a recipient of Hong Kong Postgraduate Fellowship. His current research interests include computer vision, image processing, and deep learning. He has authorized and co-authorized on more than 20 papers in top-tier academic journals and conferences.
He serves as an area chair for the conference of VISAPP. He has been serving as a reviewer for numerous academic journals and conferences such as TPAMI, TIP, TMM, TC, TNNLS, CVPR2018 and IJCAI2018.
\end{IEEEbiography}

\begin{IEEEbiography}[{\includegraphics[width=1in,height=1.25in,clip,keepaspectratio]{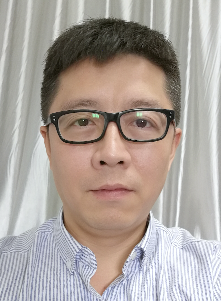}}]
{Qing Wang} is an associate professor of computer science at Sun Yat-sen University. His research focuses on human-computer interaction, user experience, collaborative software, and Web usability, with special interest in utilizing browser history in collaboration. Wang received his PhD in computer science from Sun Yat-sen University. He is a member of SIGCHI. Contact him at ericwangqing@gmail.com.
\end{IEEEbiography}

\begin{IEEEbiography}[{\includegraphics[width=1in,height=1.25in,clip,keepaspectratio]{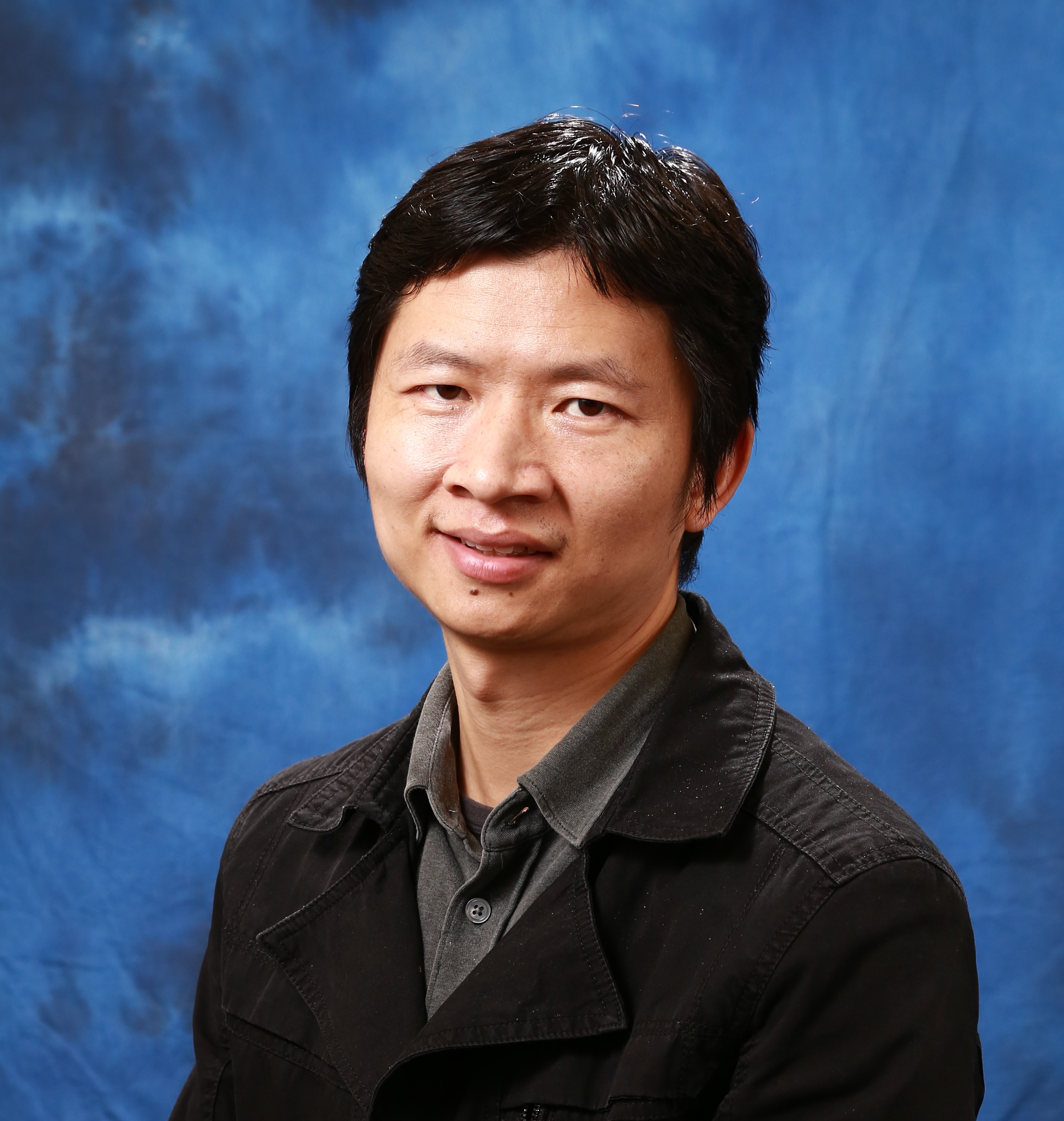}}]
{Wanli Ouyang} received the PhD degree in the Department of Electronic Engineering, The Chinese University of Hong Kong. He is now a senior lecturer in the School of Electrical and Information
Engineering at the University of Sydney, Australia. His research interests include image processing, computer vision and pattern recognition. He is a senior member of IEEE.
\end{IEEEbiography}

\begin{IEEEbiography}[{\includegraphics[width=1in,height=1.25in,clip,keepaspectratio]{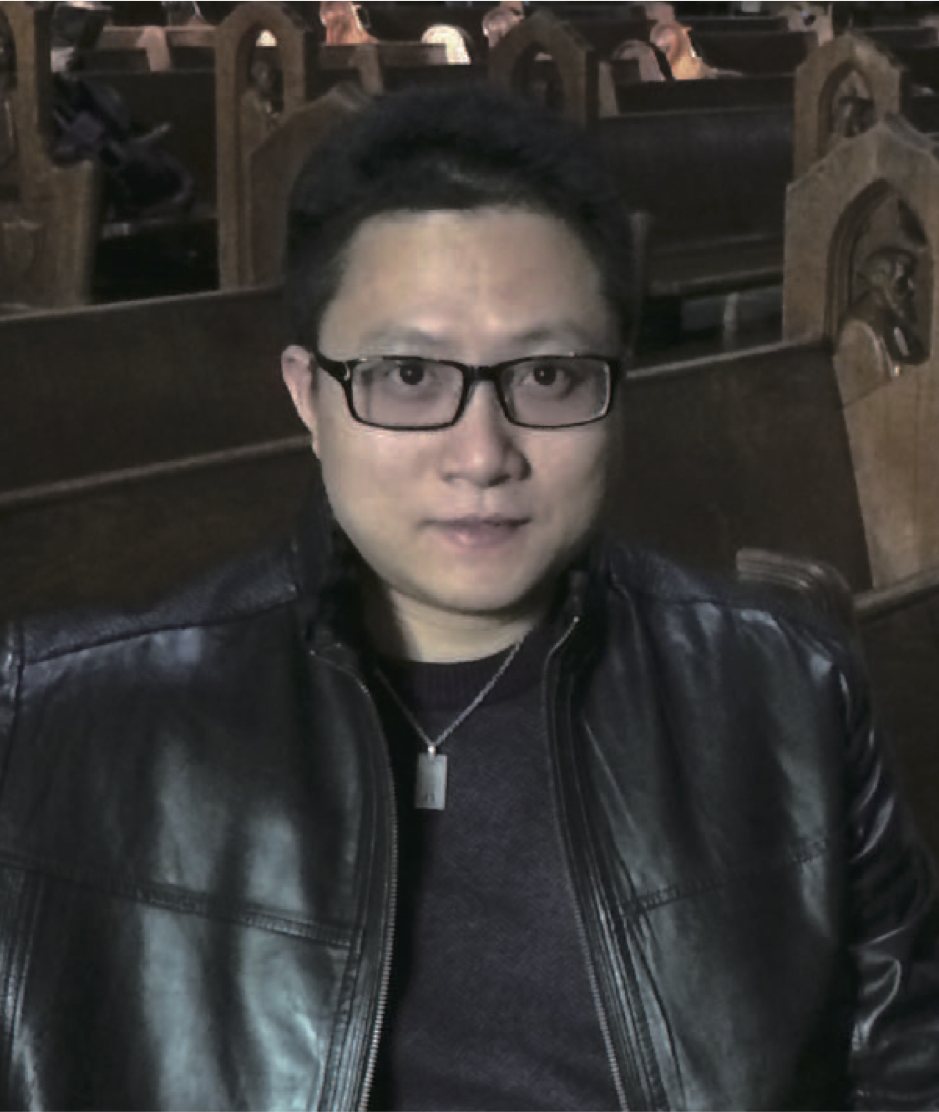}}]
{Liang Lin}
is a full Professor of Sun Yat-sen University. He is the Excellent Young Scientist of the National Natural Science Foundation of China. From 2008 to 2010, he was a Post-Doctoral Fellow at University of California, Los Angeles. From 2014 to 2015, as a senior visiting scholar, he was with The Hong Kong Polytechnic University and The Chinese University of Hong Kong. He currently leads the SenseTime R\&D teams to develop cutting-edges and deliverable solutions on computer vision, data analysis and mining, and intelligent robotic systems. He has authorized and co-authorized on more than 100 papers in top-tier academic journals and conferences. He has been serving as an associate editor of IEEE Trans. Human-Machine Systems, The Visual Computer and Neurocomputing. He served as Area/Session Chairs for numerous conferences such as ICME, ACCV, ICMR. He was the recipient of Best Paper Runners-Up Award in ACM NPAR 2010, Google Faculty Award in 2012, Best Paper Diamond Award in IEEE ICME 2017, and Hong Kong Scholars Award in 2014. He is a Fellow of IET.
\end{IEEEbiography}




\end{document}